\documentclass[lettersize,journal]{IEEEtran}
\usepackage{amsmath,amsfonts}
\usepackage{algorithmic}
\usepackage{algorithm}
\usepackage{array}
\usepackage[caption=false,font=footnotesize,labelfont=rm,textfont=rm]{subfig}
\usepackage{textcomp}
\usepackage{stfloats}
\usepackage{url}
\usepackage{verbatim}
\usepackage{graphicx}
\usepackage{cite}
\usepackage{multirow}
\usepackage{color}
\usepackage{booktabs}
\usepackage[table,xcdraw]{xcolor}
\usepackage[normalem]{ulem}
\usepackage[backref]{hyperref}
\hypersetup{hypertex=true,
	colorlinks=true,
	linkcolor=blue,
	anchorcolor=blue,
	citecolor=blue
}
\usepackage[font=footnotesize,labelfont=rm,textfont=rm]{caption}
\begin{document}
	
%%%%%%%%% TITLE
\title{TFDet: Target-Aware Fusion for RGB-T Pedestrian Detection}

\author{Xue~Zhang, Xiaohan~Zhang, Jiangtao~Wang, Jiacheng~Ying, Zehua~Sheng, Heng~Yu, Chunguang~Li, \emph{Senior Member, IEEE}, and Hui-Liang~Shen, \emph{Senior Member, IEEE}
\thanks{This work was supported in part by the National Key R\&D Program of China under grant 2023YFB3209800, in part by the Natural Science Foundation of Zhejiang Province under grant D24F020006, in part by the National Natural Science Foundation of China under grant 62301484, and in part by the Jinhua Science and Technology Bureau Project under grant 2023-1-118. \emph{(Corresponding authors: Hui-Liang Shen.)}}
\thanks{X. Zhang, X. Zhang, J. Wang, J. Ying, Z. Sheng, and C. Li are with the College of Information Science	and Electronic Engineering, Zhejiang University, Hangzhou 310027, China (e-mail: zxue2019@zju.edu.cn, zhangxh2023@zju.edu.cn, wang\_jiang\_tao\_1@163.com, yingjiacheng@zju.edu.cn, shengzehua@zju.edu.cn, cgli@zju.edu.cn).

H.-L. Shen is with the College of Information Science and Electronic Engineering, Zhejiang University, Jinhua Institute of Zhejiang University, and also with the Key Laboratory of Collaborative Sensing and Autonomous Unmanned Systems of Zhejiang Province, China (e-mail: shenhl@zju.edu.cn).
	
H. Yu is with the School of Computer Science, University of Nottingham Ningbo China, Ningbo 315199, China (e-mail: heng.yu@nottingham.edu.cn).}
}
%% The paper headers
%% Remember, if you use this you must call \IEEEpubidadjcol in the second
%% column for its text to clear the IEEEpubid mark.

\maketitle

%%%%%%%%% ABSTRACT
\begin{abstract}
	Pedestrian detection plays a critical role in computer vision as it contributes to ensuring traffic safety.
	Existing methods that rely solely on RGB images suffer from performance degradation under low-light conditions due to the lack of useful information.
	To address this issue, recent multispectral detection approaches have combined thermal images to provide complementary information and have obtained enhanced performances. 
	Nevertheless, few approaches focus on the negative effects of false positives caused by noisy fused feature maps.
	Different from them, we comprehensively analyze the impacts of false positives on the detection performance and find that enhancing feature contrast can significantly reduce these false positives.
	In this paper, we propose a novel \textbf{t}arget-aware \textbf{f}usion strategy for multispectral pedestrian \textbf{det}ection, named TFDet. 
	The target-aware fusion strategy employs a fusion-refinement paradigm. In the fusion phase, we reveal the parallel- and cross-channel similarities in RGB and thermal features, and learn an adaptive receptive field to collect useful information from both features. In the refinement phase, we use a segmentation branch to discriminate the pedestrian features from the background features. We propose a correlation-maximum loss function to enhance the contrast between the pedestrian features and background features. As a result, our fusion strategy highlights pedestrian-related features and suppresses unrelated ones, generating more discriminative fused features.
	TFDet achieves state-of-the-art performance on two multispectral pedestrian benchmarks, KAIST and LLVIP, with absolute gains of 0.65\% and 4.1\% over the previous best approaches, respectively. TFDet can easily extend to multi-class object detection scenarios. It outperforms the previous best approaches on two multispectral object detection benchmarks, FLIR and M3FD, with absolute gains of 2.2\% and 1.9\%, respectively. Importantly, TFDet has comparable inference efficiency to the previous approaches, and has remarkably good detection performance even under low-light conditions, which is a significant advancement for ensuring road safety.
	The code will be made publicly available at \url{https://github.com/XueZ-phd/TFDet.git}.
\end{abstract}

\begin{IEEEkeywords}
	RGB-T pedestrian detection, RGB-T object detection, multispectral feature fusion, feature enhancement
\end{IEEEkeywords}

%%%%%%%%% BODY TEXT
\section{Introduction}	
\begin{figure*}[t]
	\centering
	\includegraphics[width=0.7\linewidth]{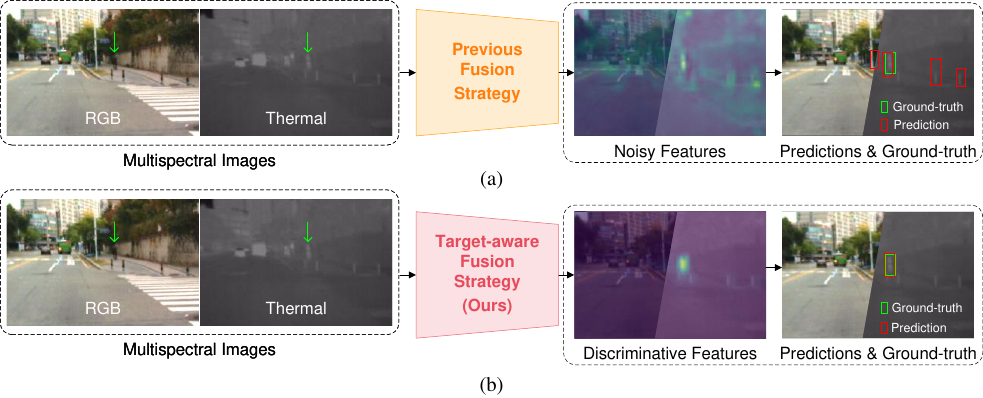}
	\caption{Visualization of features and detection results generated by a pair of multispectral images using different fusion strategies. (a) The previous fusion strategy generates noisy features and induces many false positives in the background regions. (b) Our target-aware fusion strategy generates discriminative features and effectively reduces false positives. The green arrows in the left column mark the location of pedestrians. The green boxes and red boxes in the right column indicate the ground-truth and predicted bounding boxes, respectively.}
	\label{fig:noisy features and tf features}
	\vspace{-14pt}
	\end{figure*}

	Pedestrian detection is a key problem in computer vision, with various applications such as autonomous driving and video surveillance. Modern studies mainly rely on RGB images and perform well under favorable lighting conditions. Their performance decreases under low-light conditions due to the low signal-to-noise ratio of RGB images in such scenes. In contrast, thermal images can capture the shape of the human body clearly even under poor lighting conditions. Thermal images cannot record color and texture details, limiting their ability to distinguish confusing structures. To improve the detection performance under varying lighting conditions, multispectral pedestrian detection emerges as a promising solution~\cite{kaist, llvip}. It uses the complementary information of RGB and thermal images to detect pedestrians. This complementarity is illustrated in the left column of Fig.~\ref{fig:noisy features and tf features}. When using both the RGB and thermal images, we can more easily localize the pedestrian (marked by the green arrow).
	
	%%%%%%%%%%%%
	%%%%%%%%%%%
	Recent works demonstrate that the combination of multispectral features improves the accuracy of single-modality pedestrian detection~\cite{lgfapf, msds, halffusion, dcmnet, baanet, arcnn, mbnet}. Moreover, the gain in accuracy largely depends on the fusion stage and the employed strategy. Previous studies reveal that the halfway fusion approach outperforms both early and late fusion approaches~\cite{halffusion, arcnn, LRAF-Net, cmx}. The terms ``early", ``halfway", and ``late" fusion refer to the fusion of multispectral information at the low, middle, and high stages of a two-branch network, respectively. Building upon this finding, recent works design complex halfway fusion strategies to address various challenges, such as misalignment~\cite{arcnn, uffucg, mfpt}, modality imbalance~\cite{mbnet, scdnet} and ensemble learning~\cite{proben} issues. However, a problem across these works is the generation of noisy fused features. The reason for this phenomenon is that the previous works primarily focus on combining complementary features from both modalities while neglecting to distinguish between target and non-target features. We have observed that these noisy features may lead to a significant number of false positives (FPs) and degrade performance, as shown in Fig.~\ref{fig:noisy features and tf features}~(a). A more detailed analysis is presented in Section~\ref{sec:motivation}.
	
	To mitigate the adverse effect of noisy features, in this work we propose a target-aware fusion strategy. Within the framework of this strategy, we introduce a novel multispectral feature fusion module that leverages the inherent parallel-channel and cross-channel similarities found in paired multispectral features. Additionally, we propose a feature refinement module, which includes a feature classification unit and a feature contrast enhancement unit. Furthermore, we present a novel correlation-maximum loss function. These enhancements not only enable the model to combine complementary information but also to distinguish between target and non-target features, and improve representations in target regions while suppressing those in non-target regions. Benefiting from these modules, our target-aware fusion strategy empowers the model to effectively focus on target regions, thereby reducing FPs in the background regions, as illustrated in Fig.~\ref{fig:noisy features and tf features}~(b).
	
	The detector equipped with our \textbf{t}arget-aware \textbf{f}usion strategy for multispectral pedestrian \textbf{det}ection is named TFDet. Our target-aware fusion strategy is flexible. It can be applied to both one-stage and two-stage detectors and can be easily extended to multispectral object detection scenarios that involve multiple categories such as pedestrians, cars, and other targets. TFDet achieves state-of-the-art performance on two multispectral pedestrian detection benchmarks~\cite{kaist, llvip}, as well as two multispectral object detection benchmarks~\cite{Flir, TarDAL}. More importantly, it demonstrates significantly greater superiority over previous methods under low-light conditions. Additionally, our fusion strategy is computationally efficient, enabling TFDet to generate predictions more quickly in traffic scenarios. 
	In summary, our three main contributions are:
	\begin{itemize}
		% what
		\item We propose a new multispectral feature fusion method known as the target-aware fusion strategy. Different from previous fusion strategies, our method not only focuses on combining complementary features from both modalities but also explore feature refinement techniques to enhance the representations in target regions while suppressing those in non-target regions.
		% how
		\item Our target-aware fusion strategy generates discriminative features and significantly reduces false positives in background regions. Meanwhile, our fusion strategy is flexible. It can effectively benefit both one-stage and two-stage detectors and can be extended to multi-class object detection scenarios that contain various targets.
		% result
		\item Extensive experiments demonstrate that the detector equipped with our target-aware fusion strategy, named TFDet, achieves state-of-the-art performance on four challenging datasets. Additionally, TFDet achieves comparable inference time to previous state-of-the-art counterparts, demonstrating its efficiency.
	\end{itemize}
	
	The remaining sections are organized as follows. In Section~\ref{sec:relatedwork}, we review the related works in the fields of general object detection, pedestrian detection, and multispectral pedestrian detection. In Section~\ref{sec:method}, we introduce the details of our target-aware fusion strategy and the architecture of TFDet. In Section~\ref{sec:experiments}, we present extensive experiments and analysis to evaluate the performance of our TFDet. Finally, we provide the conclusion of our work in Section~\ref{sec:conclusions}.

	%-------------------------------------------------
	\section{Related Work}
	\label{sec:relatedwork}
	In this section, we first briefly review the general object detection and pedestrian detection, and then introduce the related works of multispectral pedestrian detection.
	%-------------------------------------------------
	\subsection{Object Detection and Pedestrian Detection}
	Object detectors can be categorized into two groups: one-stage detectors~\cite{yolo, detr} and two-stage detectors~\cite{fasterrcnn, sparsercnn}, based on whether they require object proposals~\cite{detection20years}.
	One-stage detectors directly predict object boxes, either by using anchors~\cite{yolov5} or a grid of potential object centers~\cite{fcos}. YOLO~\cite{yolo} is a popular one-stage detector that takes the full-size image as input and simultaneously outputs the object boxes and class-specific confidence scores. 
	Although one-stage detectors tend to be fast, their accuracy is somehow limited. In comparison, two-stage detectors frame object detection as a ``coarse-to-fine" process. They predict object boxes based on region proposals~\cite{fasterrcnn}. In the first stage, the detector generates multiple object proposals, each indicating a potential object in an image region. In the second stage, the detector further refines the location and predicts the class scores for these proposals. Faster R-CNN~\cite{fasterrcnn} is a well-known two-stage detector. It consists of two components: RPN~\cite{fasterrcnn} and R-CNN~\cite{fastrcnn}. The RPN component is trained end-to-end to generate high-quality region proposals, which are then used by R-CNN for object detection.
	Although two-stage detectors have high accuracy, they tend to be slower.
	
	Pedestrian detection is an important application of object detection. With the advancement of object detection, pedestrian detectors can also be classified into one-stage and two-stage detectors~\cite{csppedet, generalizablepedet}. However, these pedestrian detectors rely only on RGB images, which causes their performance to degrade in low-light conditions. A promising solution to this issue is to use multispectral images.

	\subsection{Multispectral Pedestrian Detection}
	 Multispectral pedestrian detection uses RGB and thermal images to localize pedestrians~\cite{kaist}. An important research direction for this technique is how to fuse RGB and thermal data so that detectors can achieve robust detection under various illumination conditions. %There are four fusion strategies: early, halfway, late, and score fusion ~\cite{halffusion}. The first three strategies integrate features of RGB and thermal images at the early, middle, and final levels of a detector, respectively~\cite{EME, M2FNet, dcmnet}. The score fusion strategy averages the scores of the predicted boxes at the same location in RGB and thermal images~\cite{halffusion, arcnn, proben}. Among these fusion strategies, the halfway fusion strategy typically achieves superior detection performance~\cite{halffusion, msds, lgfapf}. Early approaches using the halfway fusion strategy simply concatenate or add multispectral features~\cite{halffusion, msds}, without adaptively adjusting the importance of features from different modalities in various environments. 
	 	
 	For the adaptive feature fusion purpose, recent works develop illumination-aware feature fusion approaches~\cite{illuminationaware, RITA, adaptation, IGT, TINet}, attention based feature fusion approaches~\cite{TIRDet, M2FNet, CAT+MFT, MSAT, ARCNN1, C2Former}, and non-local feature fusion approaches~\cite{dcmnet}. The illumination-aware feature fusion approaches typically introduce a classification branch to determine the importance of RGB features based on the illumination conditions. Considering that the classification result of an image cannot reflect the importance of individual region, attention based approaches use spatial attention, channel attention, or cross-attention of transformer to assist feature fusion~\cite{TIRDet, cmx, M2FNet, CAT+MFT, C2Former, mfpt}. Spatial attention and channel attention can generate element-wise and channel-wise weighting factors for multispectral features, while cross-attention can model global contextual correlations. Cross-attention has the ability of addressing the misalignment problem between the multispectral features, but it is computationally expensive. This problem can be alleviated by non-local feature aggregation~\cite{dcmnet}. Different from previous approaches, we propose a modified deformable convolution to explicitly model the offsets between the RGB and thermal features. We also reveal the inherent parallel- and cross-channel similarities in multispectral features and propose a two-step method for adaptive feature fusion.
	 	
 	With the success of weakly supervised learning in general object detection~\cite{zhang2021weakly}, recent multispectral pedestrian detection approaches employ box-level masks to improve detection performance~\cite{msds, gaff, lgfapf, M2FNet}. The box-level mask is a binary mask with the pedestrian regions filled with 1s and other regions with 0s. MSDS-RCNN~\cite{msds} introduces a segmentation branch and utilizes the binary mask to supervise its training. GAFF~\cite{gaff} uses the box-level mask to guide the attention of inter- and intra-modality. As these approaches don't focus on the globally information from pedestrians, LG-FAPF~\cite{lgfapf} and M2FNet~\cite{M2FNet} utilize the box-level mask and the cross-attention of transformer to enhance the global modeling ability. We note that previous approaches neglect the impact of noisy feature maps on detection performance, while the noisy fused features may cause false positives in background regions. In this work, we focus on the negative impact of false positives on detection performance and use the box-level mask to enhance the feature contrast between pedestrian and background regions.
	%-------------------------------------------------
	\section{Method}\label{sec:method}
	In this section, we first describe our motivation. Then, we present the mathematical principles behind our proposed target-aware fusion strategy in Section~\ref{sec:model formulation}. Subsequently, we provide an overview of TFDet in Section~\ref{sec:overview} and elaborate on its three key components: the feature fusion module (FFM) in Section~\ref{sec:ffm}, the feature refinement module (FRM) in Section~\ref{sec:frm}, and the correlation-maximum loss function in Section~\ref{sec:loss}.
	
	We are also interested in understanding the practical operational mechanisms of each module in our target-aware fusion strategy. However, a rigorous theoretical analysis of the representations learned by deep neural networks remains a challenging task. Therefore, we take an empirical approach to explore the role of each module. We describe the effect of FFM immediately after introducing the structure of FFM in Section~\ref{sec:ffm}. Since the FRM and the correlation-maximum loss function are complementary for contrast enhancement during training, we describe their effects in Section~\ref{sec:loss}.

	\subsection{Motivation}
	\label{sec:motivation}
	Our motivation arises from the observation that the detector can recall most of the ground-truth pedestrians but also generates many FPs in the background regions.	As shown in Fig.~\ref{fig:noisy features and tf features}~(a), the detector makes a correct prediction in the target region but generates three FPs in background regions. Nevertheless, previous works overlook the impact of FPs on detection performance.
	%------------------------------
	% MR vs FPs
	\begin{figure}
	\begin{minipage}{\linewidth}
	\centering
	\includegraphics[width=\linewidth]{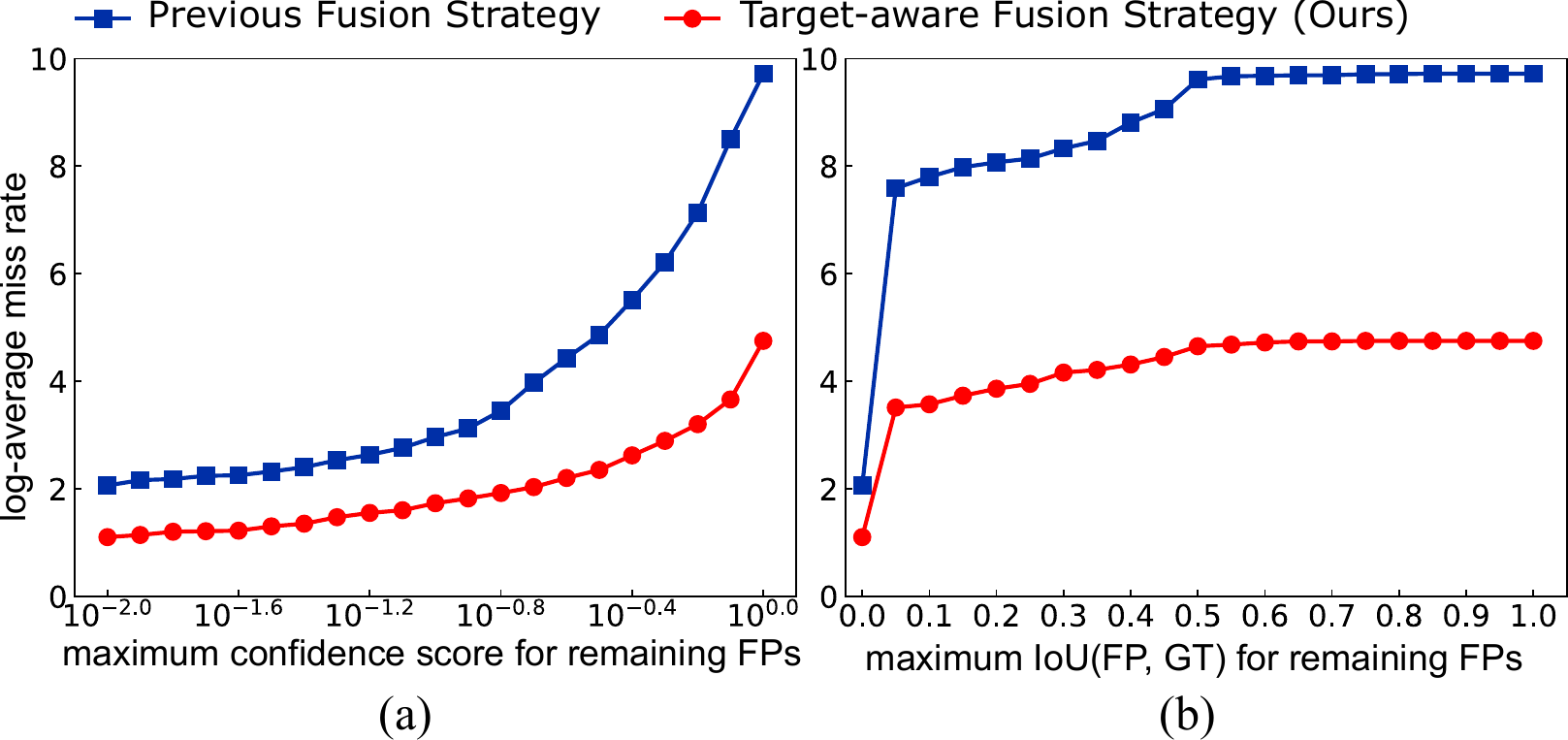}
	\caption{A pilot study for analyzing the impact of FPs on detection performance. From right to left, the detector's performance gradually improves as FPs are removed based on their (a) confidence scores and (b) IoU ratios with ground-truth boxes.}
	\label{fig:performance vs fps}
	\end{minipage}
	\begin{minipage}{\linewidth}
	\centering
	\includegraphics[width=\linewidth]{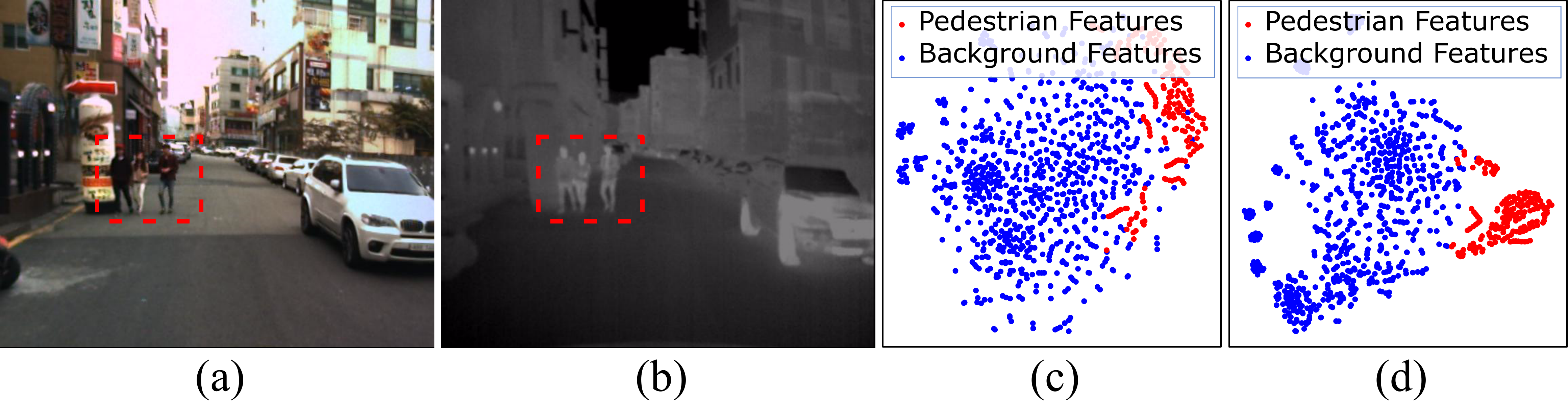}
	\caption{Visualization of pedestrian and background features in paired multispectral images using the t-SNE technique. The pedestrian region is highlighted by a red dashed box in (a) RGB image and (b) thermal image. The features generated by the previous fusion strategy and our target-aware fusion strategy are presented in (c) and (d), respectively.}
	\label{fig:tsne}
	\end{minipage}
	\vspace{-14pt}
	\end{figure}
	%------------------------------

	To analyze this impact, we conduct a pilot study. Specifically, we manually remove the FPs based on their (a) confidence scores and (b) intersection-over-union (IoU) rates with ground-truth boxes in descending order. The corresponding results are presented in Fig.~\ref{fig:performance vs fps}. The log-average miss rate progressively decreases (lower values are better) as the FPs are removed. Furthermore, the log-average miss rate decreases rapidly when FPs with (a) high scores and (b) low IoU rates are removed. This reveals that FPs with high scores may be mainly distributed in the background regions, and these FPs are more detrimental. This phenomenon may be attributed to the similarity in shape between objects in the FPs' regions and pedestrians, as shown in Fig.~\ref{fig:noisy features and tf features}~(a). The shape similarity confuses the model, leading to the generation of noisy features and a number of FPs.
	
	We propose a target-aware fusion strategy to address the noisy feature issue. We introduce a feature fusion module (FFM) to combine complementary features, a feature refinement module (FRM) to discriminate between target and non-target features, and a correlation-maximum loss function to enhance feature contrast.
	Our target-aware fusion strategy significantly reduces FPs, as shown in Fig.~\ref{fig:performance vs fps}. Additionally, we showcase the feature visualizations of both the previous fusion strategy and our target-aware fusion strategy using t-distributed stochastic neighbor embedding (t-SNE)~\cite{tsne}, as depicted in Fig.~\ref{fig:tsne}. Within the t-SNE map, our target-aware fusion strategy tightly clusters points of the same class and effectively separates those from different classes. This demonstrates that our fusion strategy can generate contrast-enhanced features.
	
	\subsection{Foundation of Feature Contrast Enhancement}\label{sec:model formulation}
\begin{figure}[t]
	\centering
	\includegraphics[width=0.8\linewidth]{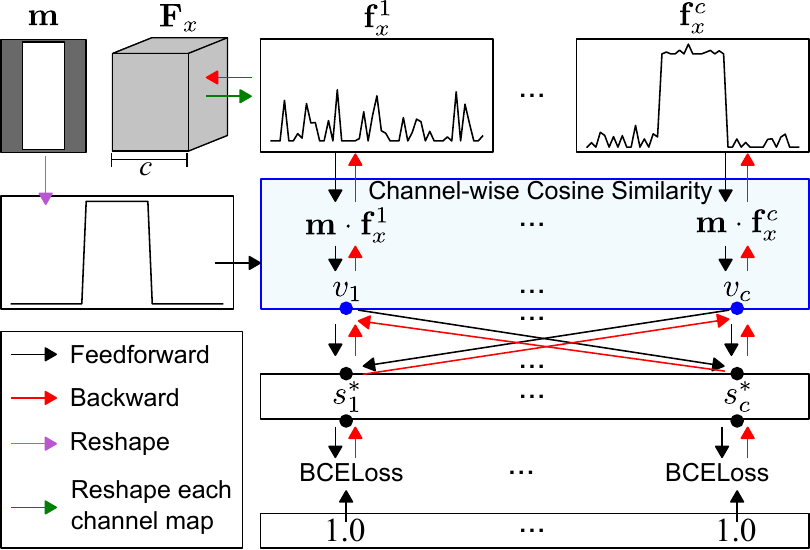}\\
	\caption{Schematic diagram of the feature contrast enhancement foundation. In the mask $\mathbf{m}$, the white region denotes the foreground region, while the dark region denotes the background region. In the backward process, the gradients are only used to update the initial fused feature $\mathbf{F}_x$. Note that in this figure $\mathbf{m}$, $\mathbf{f}^1_x$, and $\mathbf{f}^c_x$ are reshaped to vectors for illustration.}
	\label{fig:foundation}
	\vspace{-16pt}
 	\end{figure}

 	The mathematical foundation of the feature contrast enhancement in our fusion strategy is built upon the correlation between a ground-truth (GT) box-level mask and the initial fused feature. The initial fused feature is generated by a feature fusion module. This feature contains full information from both modalities, but it is noisy. The box-level mask is generated by assigning 1s to the regions inside GT boxes and 0s to the other regions, hence it has a distinct contrast between the foreground regions and the background regions. In this context, the intuition for contrast enhancement of the initial fused feature is to improve the aforementioned correlation in the channel dimension. Fig.~\ref{fig:foundation} illustrates the schematic diagram of the feature contrast enhancement foundation. In the following we elaborate the computation process.
 	
	%------------------foundation----------
 	%------------------------------------
 	
 	%-----------------overview-----------
 	\begin{figure*}[t]
	\centering
	\includegraphics[width=0.9\linewidth]{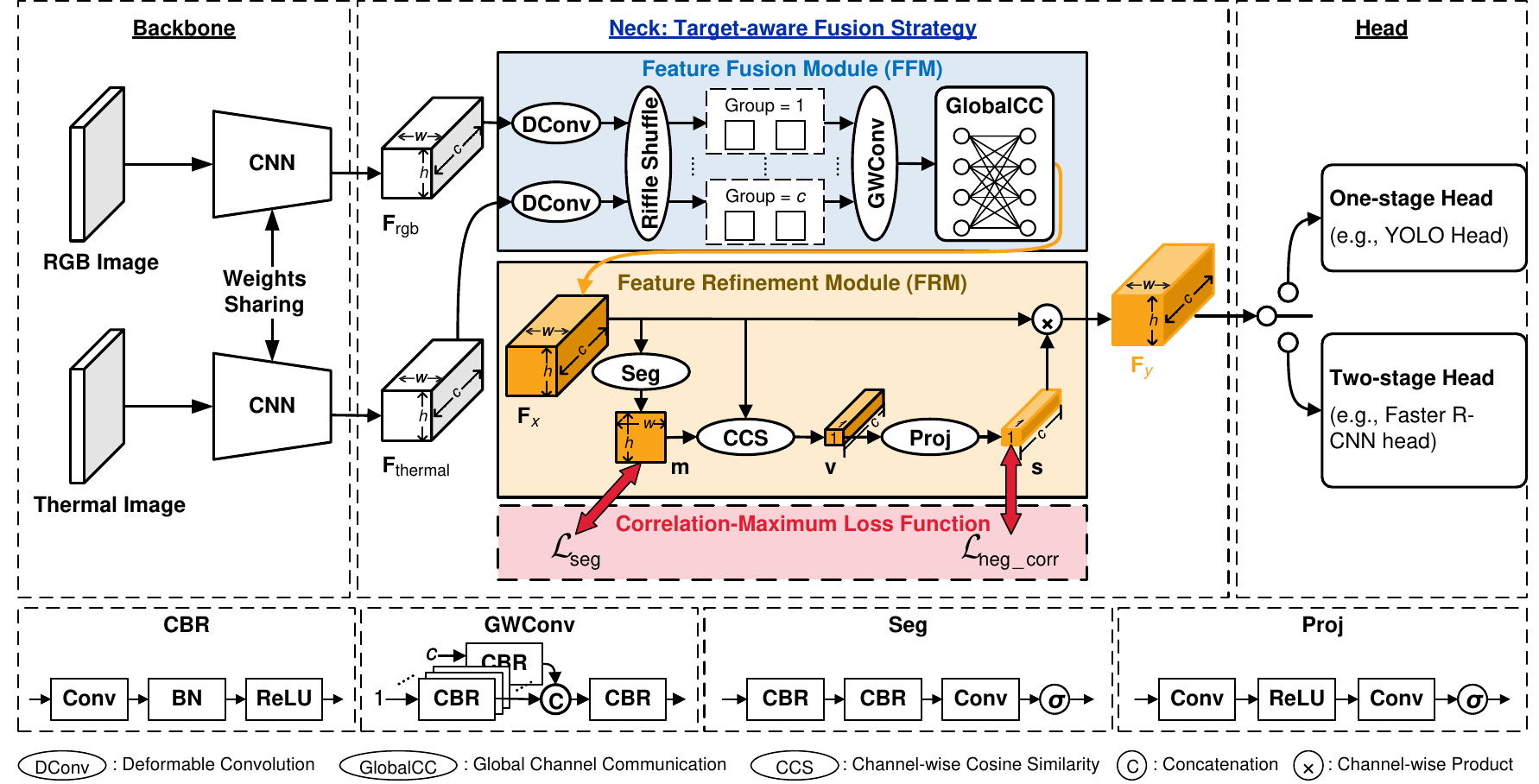}
	\caption{The illustration of our TFDet architecture. It consists of three components: backbone, neck, and head. The backbone extracts features from paired multispectral images. The neck fuses these multispectral features using our target-aware fusion strategy. The head generates boxes and corresponding scores based on the fused feature. A more detailed structure of the feature fusion module (FFM) is shown in Fig.~\ref{fig:ffm}. The correlation-maximum loss function is only used during the training phase of our model.}
	\label{fig:overview}
	\vspace{-14pt}
 	\end{figure*}
 	%----------------------------
	
	Denote the initial fused feature as $\mathbf{F}_x\in\mathbb{R}^{h\times w\times c}$ and the GT box-level mask as $\mathbf{\tilde{m}}\in \mathbb{R}^{h\times w}$. Since the GT mask $\mathbf{\tilde{m}}$ is not available at test time, we learn to predict it using a segmentation branch. Denote the predicted mask as $\mathbf{m}\in \mathbb{R}^{h\times w}$, we can compute the cosine similarity between the predicted mask and the initial fused feature across the channel dimension as
	\begin{equation}\label{equ:correlation_v}
	\begin{aligned}
	\mathbf{v}&=\left[v_1, v_2, \dots, v_c\right]\\
	&=\left[\mathbf{m}\cdot\mathbf{f}_x^1, \mathbf{m}\cdot\mathbf{f}_x^2,\dots,\mathbf{m}\cdot\mathbf{f}_x^c\right],
	\end{aligned}
	\end{equation}
	where the operator ``.'' denotes the matrix inner product. $\mathbf{f}_x^i\in\mathbb{R}^{h\times w}$$(1\le i \le c)$ is the $i$-th channel feature of $\mathbf{F}_x$. Since both $\mathbf{m}$ and $\mathbf{f}_x^i$ are normalized, the inner product can indicate their similarities.
	
	One straightforward method for enhancing feature contrast is to maximize each entry in the correlation $\mathbf{v}$. However, this method makes each feature map equal to the box-level mask, resulting in the loss of important semantic information across different channels. To address this issue, we introduce a projection matrix $\mathbf{w}\in\mathbb{R}^{c\times c}$ to encode the correlation $\mathbf{v}$ and generate the processed correlation
	\begin{equation}
		\mathbf{s^*} =\left[s^*_1, s^*_2, \dots, s^*_c\right]^\mathsf{T}=\mathbf{w}\mathbf{v}^\mathsf{T},
	\end{equation}
	where the superscript ``$^\mathsf{T}$'' represents the transpose operator.
	
	In this context, an entry of the processed correlation $\mathbf{s^*}$ can be written as 
	\begin{equation}
		\begin{aligned}
			s^*_i &=\sum_{j=1}^{c}\mathbf{w}(i, j)v_j\\
			&=\mathbf{m}\cdot\sum_{j=1}^{c}\mathbf{w}(i, j)\mathbf{f}_x^j.
		\end{aligned}
	\end{equation}
	Let $\left\{s^*_i\rightarrow1|i\in[1,c]\right\}$, we can infer that 
	\begin{equation}
		\forall i\in[1,c], \sum_{j=1}^{c}\mathbf{w}(i, j)\mathbf{f}_x^j\rightarrow\mathbf{m},
	\end{equation}
	where $\mathbf{w}(i, j)$ denotes the element located at row $i$ and column $j$ within the matrix $\mathbf{w}$.
	This equation demonstrates that each entry $s^*_i$ represents the similarity between the predicted mask and the weighted average of initial fused feature across the channel dimension. Based on this analysis, we enforce each entry $s^*_i$ to approximate 1.0, thereby enhancing their correlation. The mathematical process described above is linear, thus we use convolutional operations to implement the projection process in our target-aware fusion strategy. In addition, we introduce a correlation-maximum loss function, which includes a segmentation loss function and a negative correlation loss function, to supervise the predicted mask and the processed correlation.

	\subsection{Architecture of TFDet}
	\label{sec:overview}
	The overall architecture of our TFDet is illustrated in Fig.~\ref{fig:overview}. We feed a pair of RGB and thermal images into a CNN backbone network to generate the corresponding multispectral features denoted as $\mathbf{F}_{\rm rgb}$ and $\mathbf{F}_{\rm thermal}$. Subsequently, the multispectral features are fused in the neck part using our target-aware fusion strategy. In this fusion strategy, the FFM first adaptively combines the multispectral features, generating the initial fused feature denoted as $\mathbf{F}_x$. Based on this initial fused feature, the FRM then discriminates between the foreground and background features by employing an auxiliary segmentation branch. Following that, the FRM enhances feature contrast through the correlation-maximum loss function during training. Our target-aware fusion strategy finally yields the discriminative feature $\mathbf{F}_y$, which is subsequently fed into detection heads. These detection heads can be the one-stage head (such as YOLO head~\cite{yolov5}) or the two-stage head (such as Faster R-CNN head~\cite{fasterrcnn}, including the RPN and R-CNN). This process predicts boxes and corresponding scores.
	
	Our target-aware fusion strategy is independent of the CNN structure. Different detectors often leverage various pyramid networks to fuse multi-scale features. For instance, Faster R-CNN employs FPN~\cite{fpn}, while YOLOv5 utilizes PANet~\cite{panet}. Due to this, we omit the schematic diagram of the pyramid networks in Fig.~\ref{fig:overview}. Since we conduct the identical feature fusion process at different levels of the pyramid networks, we only provide a detailed description of the process at one level.
		
	In this paper, we employ a weight-sharing CNN as the backbone network to extract multispectral features. This choice is motivated by its ability to save computation costs and the reduction in learnable parameters, which helps mitigate the potential effects of overfitting. Meanwhile, we also notice that certain works~\cite{TINet, dfanet, scdnet, mbnet, lgfapf, gaff} use different backbone networks for feature extraction. Nevertheless, such approaches often need to elaborately design complex training techniques to ensure the model's effective generalization~\cite{dfanet}.
	
	%-------------------------------------------------
	\subsection{Feature Fusion Module (FFM)}
	\label{sec:ffm}
	%%%%%%%%%%%%%%%%%%%%%%%%%%%%%%%%%%%%%%%%%%%
	\begin{figure}[t]
	\centering
	\subfloat[]{\includegraphics[width=0.61\linewidth]{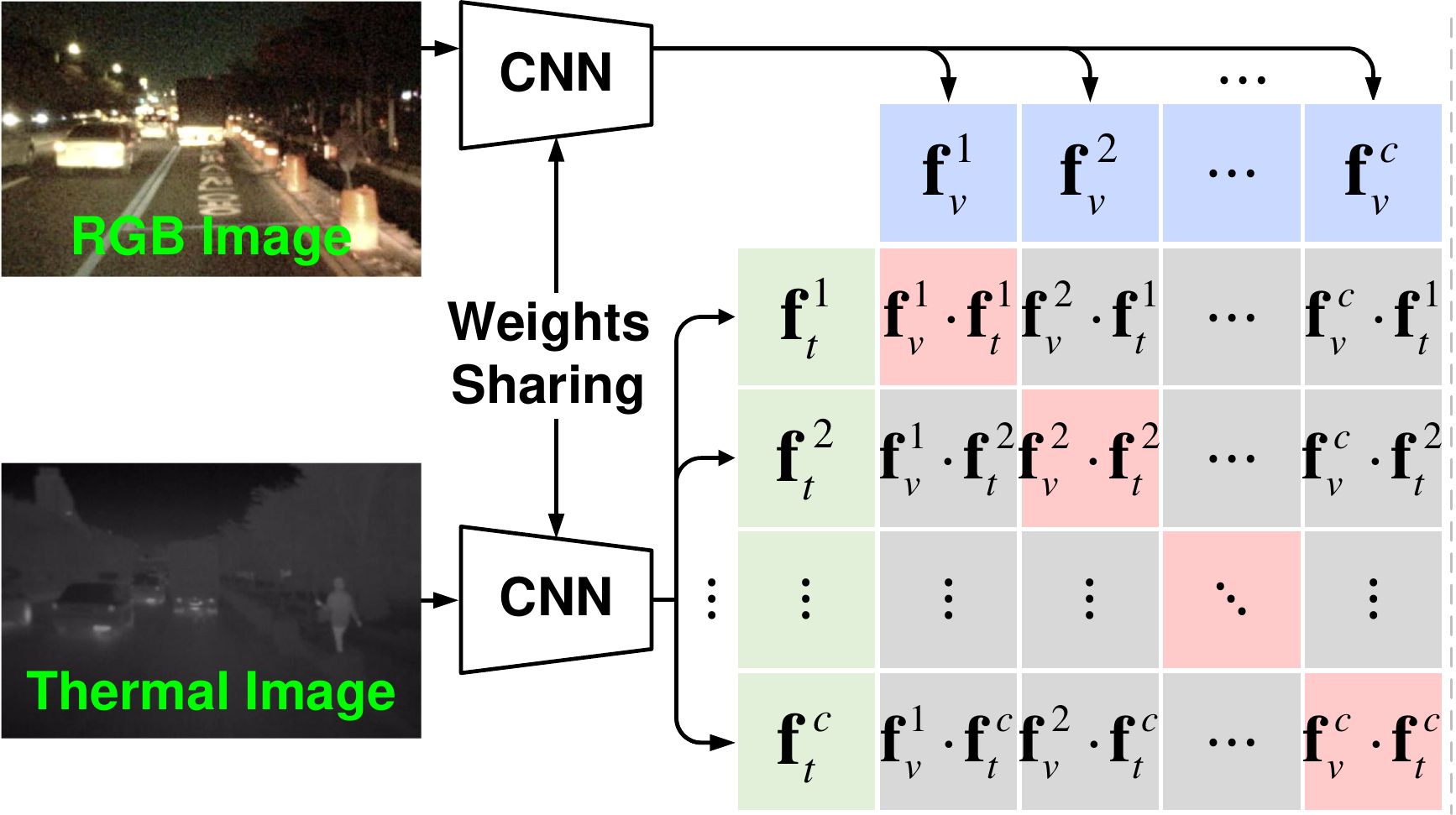}}\hspace{-3pt}
	\subfloat[]{\includegraphics[width=0.385\linewidth]{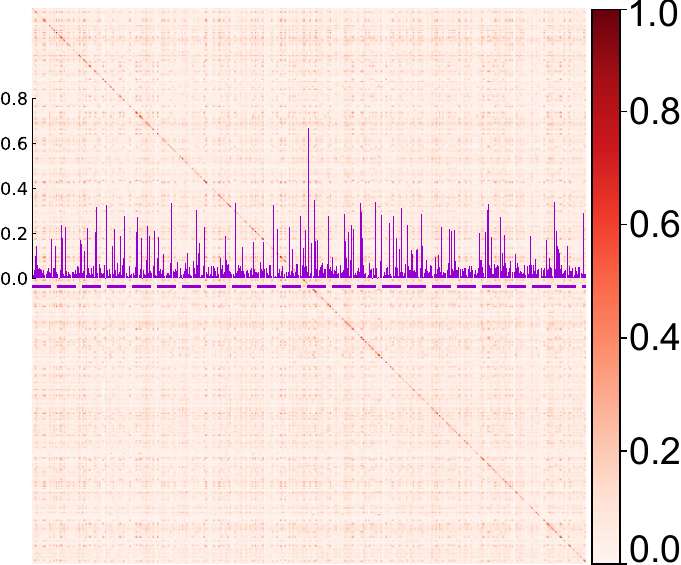}}
	\caption{Feature relation matrix of paired multispectral images. (a) The mathematical process of computing the relation matrix. (b) The resulting relation matrix. In (b), the dashed purple line marks the middle row of the matrix, and the 1-D distribution of this row is shown in purple.}
	\label{fig:feature_similarities}
	\vspace{-18pt}
	\end{figure}
	%%%%%%%%%%%%%%%%%%%%%%%%%%%%%%%%%%%%%%%%%%%
	\textbf{Observation.}
	Our proposed FFM gains insights from the similarities in multispectral features. Specifically, given the multispectral features $\mathbf{F}_{\rm rgb}\in\mathbb{R}^{h\times w \times c}$ and $\mathbf{F}_{\rm thermal}\in\mathbb{R}^{h\times w \times c}$, each of which contains $c$ channel maps with a resolution of $h \times w$, we exhaustively compute the matrix inner product among all channel maps, obtaining a multispectral feature relation matrix of size $c  \times c$. This process is depicted in Fig.~\ref{fig:feature_similarities}~(a), where we denote the features $\mathbf{F}_{\rm rgb}$ and $\mathbf{F}_{\rm thermal}$ as $\left\{\mathbf{f}_v^{i}|\mathbf{f}_v^{i}\in\mathbb{R}^{h\times w}, i\in[1,c]\right\}$ and $\left\{\mathbf{f}_t^{i}|\mathbf{f}_t^{i}\in\mathbb{R}^{h\times w}, i\in[1,c]\right\}$, respectively. From Fig.~\ref{fig:feature_similarities}~(b), we can observe two enlightening phenomena: (1) elements along the diagonal have larger values, indicating a strong similarity between multispectral features at the same channel position, which we refer to as ``\textbf{parallel-channel similarity}'', and (2) there are also large elements in the off-diagonal regions, suggesting that features across different channels may also exhibit strong similarity, which we term as ``\textbf{cross-channel similarity}''. 
	
	To comprehensively verify whether the similarities in multispectral features are universal, we define two metrics: the \textbf{A}verage Mea\textbf{n} \textbf{R}atio (ANR) and the \textbf{A}verage Med\textbf{i}an \textbf{R}atio (AIR). These metrics are designed to summarize the differences between diagonal and off-diagonal elements. The ANR computes the average ratio between the element at the diagonal position and the mean of those at the off-diagonal positions across the entire dataset. Meanwhile, the AIR employs a similar approach but utilizes the median value of the off-diagonal elements as the denominator. The mathematical process can be formulated as
\begin{equation}
	\begin{aligned}
		&\textrm{NR}_{n} = \frac{1}{c}\sum_{i=1}^{c}\frac{\mathbf{f}_v^i\cdot\mathbf{f}_t^i}{\textrm{Mean}\left(\left\{\mathbf{f}_v^i\cdot\mathbf{f}_t^j | j \in \left[1,c\right] \backslash \{i\} \right\}\right)},\\
		&\textrm{IR}_{n} = \frac{1}{c}\sum_{i=1}^{c}\frac{\mathbf{f}_v^i\cdot\mathbf{f}_t^i}{\textrm{Median}\left(\left\{\mathbf{f}_v^i\cdot\mathbf{f}_t^j | j \in \left[1,c\right] \backslash \{i\} \right\}\right)},\\
	\end{aligned}
\end{equation}
and
\begin{equation}
	\begin{aligned}
		\textrm{ANR} = \frac{1}{N}\sum_{n=1}^{N}\textrm{NR}_n,
		\textrm{AIR} = \frac{1}{N}\sum_{n=1}^{N}\textrm{IR}_n,\\
	\end{aligned}
\end{equation}
	where $N$ represents the number of paired RGB-T images in a dataset, and the entry $n$ denotes the index of the image pair. For each image pair, $\mathbf{f}_v^i\cdot\mathbf{f}_t^i$ computes the matrix inner product of multispectral features at channel position $i$. The operators Mean($\cdot$) and Median($\cdot$) return the mean and median values of the given sets, respectively. We select the mean and median values as denominators for two reasons: (1) the mean value reflects the overall expectation of elements at off-diagonal positions, and (2) the median value avoids the influence of outliers.
	
	%%%%%%%%%%%%%%%%%%%%%%%%%%%%%%%%%%%%%
	\begin{table}[t]
	\setlength{\tabcolsep}{16pt}
	\renewcommand{\arraystretch}{1.1}
	\caption{The ANR and AIR on two datasets. `\# channels' represents the number of channels in multispectral features.}% As the images are down-sampled, the number of channels increases. Specifically, when the number of channels is 512 and 1024, the images are down-sampled by a factor of 8 and 16, respectively.}
	\vspace{-12pt}
	\begin{center}
	\resizebox{\linewidth}{!}{
	\begin{tabular}{lcccc}
	\hline\hline
	{Dataset}&\multicolumn{2}{c}{KAIST~\cite{kaist}}&\multicolumn{2}{c}{LLVIP~\cite{llvip}}\\
	\cmidrule(r){1-1}\cmidrule(r){2-3} \cmidrule(r){4-5}
	{$\#$ Channels} &{512}&{1024}&{512}&{1024}\\\hline
	{ANR}&{3.52}&{1.60}&{1.48}&{1.46}\\
	{AIR}&{8.65}&{2.95}&{1.76}&{1.97}\\
	\hline\hline
	\end{tabular}}
	\end{center}
	\label{tbl:ANR_AIR}
	\vspace{-14pt}
	\end{table}
	%%%%%%%%%%%%%%%%%%%%%%%%%%%%%%%%%%%%%
	
	Based on the aforementioned definition, we calculate ANR and AIR on two representative datasets: KAIST~\cite{kaist} and LLVIP~\cite{llvip}. Detailed statistics for these two datasets are provided in Section~\ref{sec:dataset}. The results presented in Table~\ref{tbl:ANR_AIR} reveal two phenomena: (1) both the ANR and AIR exceed 1, indicating strong parallel-channel similarity, and (2) ANR consistently tends to be smaller than AIR, implying a strong cross-channel similarity.

	%%%%%%%%%%%%%%%%%%%%%%%%%%%%%%%%%%%%%%%%%%%
	\begin{figure*}[h]
	\centering
	\includegraphics[width=0.9\linewidth]{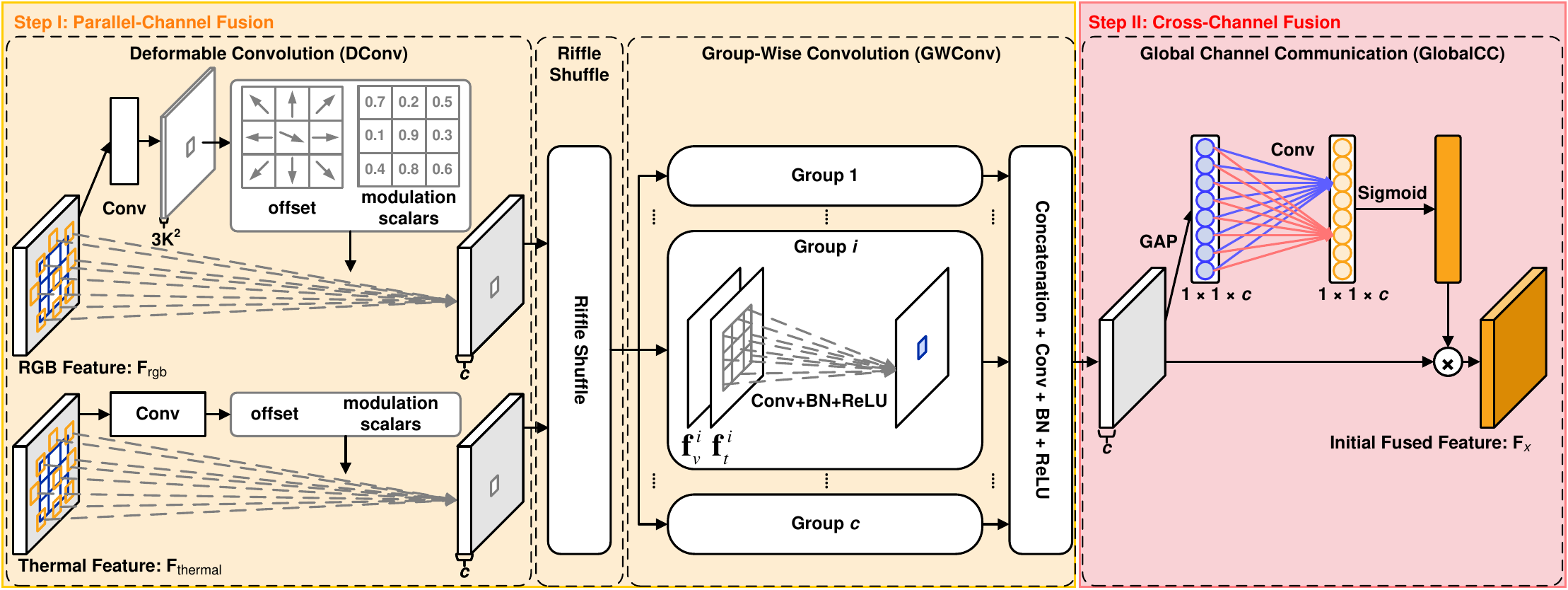}
	\caption{Illustration of the feature fusion module (FFM). This module is implemented in two steps: parallel-channel fusion and cross-channel fusion. It consists of four components. DConv adaptively learns the receptive field according to pedestrian scales. Riffle Shuffle groups two multispectral feature maps at the same channel position, thus generating $c$ groups of features. GWConv fuses the grouped features. GlobalCC performs recalibration using a self-gating mechanism.}
	\label{fig:ffm}
	\vspace{-12pt}
	\end{figure*}
	%%%%%%%%%%%%%%%%%%%%%%%%%%%%%%%%%%%%%%%%%%%
	
	\textbf{Feature Fusion Module Structure.} Based on the universal parallel- and cross-channel similarities in multispectral features, we propose a two-step method for fusing these features, as illustrated in Fig.~\ref{fig:ffm}. In the first step, we employ the property of parallel-channel similarity to fuse two corresponding channel features. Considering the diverse scales of pedestrians at varying distances from cameras, we adopt deformable convolution (DConv)~\cite{dcnv2} to adaptively learn receptive field for each output response. However, since the DConv is originally designed for single-modality inputs, we propose a modification involving riffle shuffle and group-wise convolution (GWConv) to tailor it for multispectral features. In the second step, we exploit the property of cross-channel similarity to recalibrate channel-wise features. For this purpose, we introduce a global channel communication (GlobalCC) block, which explicitly constructs interdependencies between channels, producing a set of modulation weights. These weights are then applied to the feature obtained in Step I, yielding the initial fused feature $\mathbf{F}_x$.
	
	%%%%%%%%%%%%%%%%%%%%%%%%%%%%%%%%%%%%%%%%%%%%%%%%%
	\begin{figure}[t]
	\centering
	\includegraphics[width=\linewidth]{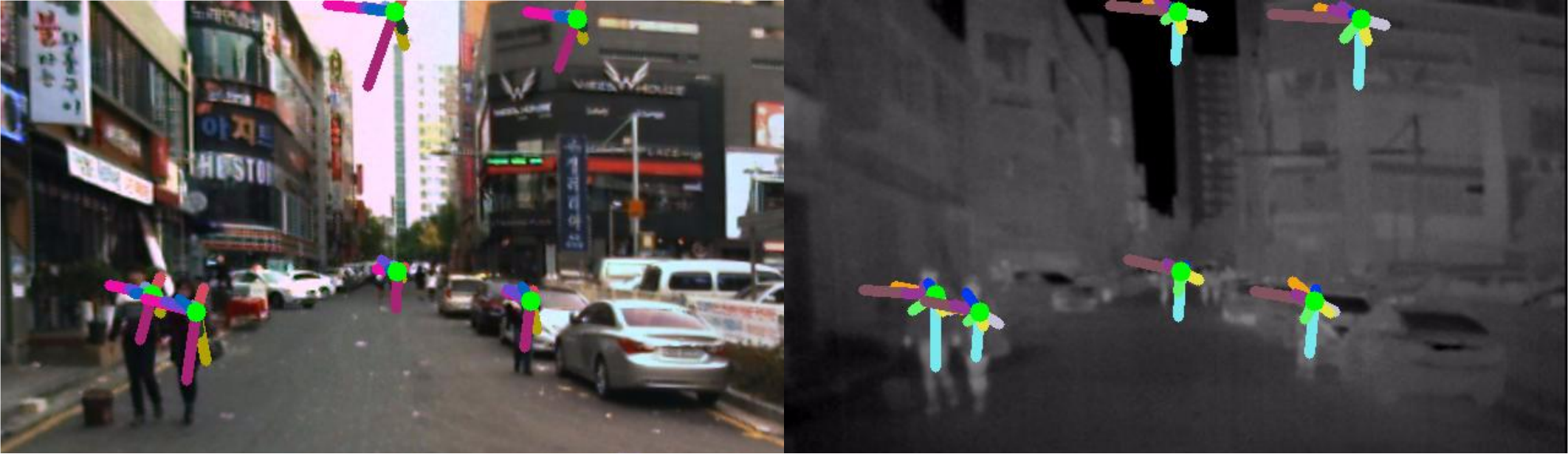}
	\caption{Sampling locations of several activation units obtained from our modified DConv. Each green point represents an activation unit, and the 9 lines surrounding the point indicate the associated sampling locations.}
	\label{fig:dcnPoints}
	\vspace{-12pt}
	\end{figure}
	%%%%%%%%%%%%%%%%%%%%%%%%%%%%%%%%%%%%%%%%%%%%%%
	\begin{table}
	\setlength{\tabcolsep}{27pt}
	\renewcommand{\arraystretch}{1.1}
	\caption{Comparison of MR (\%) between our FFM and three variants.}
	\vspace{-12pt}
	\begin{center}
	\resizebox{\linewidth}{!}{
	\begin{tabular}{lc}
	\hline\hline
	Method&MR ($\downarrow$)\\
	\hline
	Fixed-RP&{9.90}\\\hline
	Adaptive-RP&{8.96}\\\hline
	Adaptive-RP + ConvCC&{8.94}\\\hline
	FFM (Adaptive-RP + GlobalCC)&{8.57}\\
	\hline\hline
	\end{tabular}}
	\end{center}
	\label{tbl:roleofFFM}
	\vspace{-16pt}
	\end{table}
	%%%%%%%%%%%%%%%%%%%%%%%%%%%%%%%%%%%%%%%%%%%%%%%%%
	
	\textbf{Effect Validation: Adaptive Receptive Field (Adaptive-RP).} We evaluate our modified DConv's ability in adaptively learning receptive fields. In Fig.~\ref{fig:dcnPoints}, we visualize how sampling locations change with object variations, indicating adaptive receptive field behavior. We also compare Adaptive-RP with a fixed receptive field variant called Fixed-RP in Table~\ref{tbl:roleofFFM}, which uses two convolutional operations to replace the DConv. This comparison demonstrates the superiority of our Adaptive-RP.
	
	\textbf{Effect Validation: Effect of Global Chanel Communication.} To assess the impact of GlobalCC, we replace it with a $1\times 1$ convolution called ConvCC. The results in Table~\ref{tbl:roleofFFM} show that our FFM, which incorporates Adaptive-RP and GlobalCC, achieves superior performance due to GlobalCC's capacity to capture a global receptive field via global average pooling (GAP), unlike ConvCC with its fixed-sized kernel.
	%-------------------------------------------------
	\subsection{Feature Refinement Module (FRM)}
	\label{sec:frm}
	We propose FRM to enhance the contrast between foreground features and background features. FRM takes the initially fused feature $\mathbf{F}_x \in \mathbb{R}^{h\times w\times c}$ as input, and outputs the contrast-enhanced feature $\mathbf{F}_y \in \mathbb{R}^{h\times w\times c}$. Specifically, we first generate the box-level mask label based on the bounding-box annotations by filling foreground areas with 1s and the remaining areas with 0s. Next, we use the feature $ \mathbf{F}_x $ to predict the box-level mask through a segmentation (Seg) branch. This process is defined as 
	\begin{equation}
		\mathbf{m} = \sigma\left(\mathcal{H}\left(\mathbf{F}_x\right)\right),
	\end{equation}
	where the function $\mathcal{H}(\cdot)$ denotes the segmentation branch, and the function $\sigma(\cdot)$ represents a sigmoid activation.
	
	We then compute the correlation $\mathbf{v}$ between the predicted box-level mask $\mathbf{m}$ and the feature $ \mathbf{F}_x $ along the channel direction, which is defined in (\ref{equ:correlation_v}).	This correlation is processed by a projection (Proj) layer to generate the processed correlation $\mathbf{s}$
	\begin{equation}
		\mathbf{s} =\sigma\left(\mathcal{P}\left(\mathbf{v}\right)\right),
	\end{equation}
	where the function $\mathcal{P}(\cdot)$ represents the projection layer, which consists of two convolutional operations. We then use it to scale the feature $ \mathbf{F}_x $ and generate the refined feature 
	\begin{equation}
		\mathbf{F}_y = \mathbf{s} \otimes \mathbf{F}_x,
	\end{equation}
	where the operation $\otimes$ performs channel-wise product.
	Throughout the refinement process, we supervise the predicted box-level mask $ \mathbf{m} \in\mathbb{R}^{h\times w}$ and maximize the processed correlation $\mathbf{s}\in\mathbb{R}^{1\times1\times c}$ using our correlation-maximum loss function. We note that FRM can easily be extended to multi-class object detection scenarios. In such a scenario, we fill all target regions with 1s and the background regions with 0s to construct the box-level mask label, and then use the above process to enhance feature contrast. We have evaluated its extensibility in Section~\ref{sec:extension}.
		
	%-------------------------------------------------
	\subsection{Correlation-Maximum Loss Function}
	\label{sec:loss}
	%%%%%%%%%%%%%%%%%%%%%%%%%%%%%%%%
	\begin{figure*}[t]
		\centering
		\includegraphics[width=\linewidth]{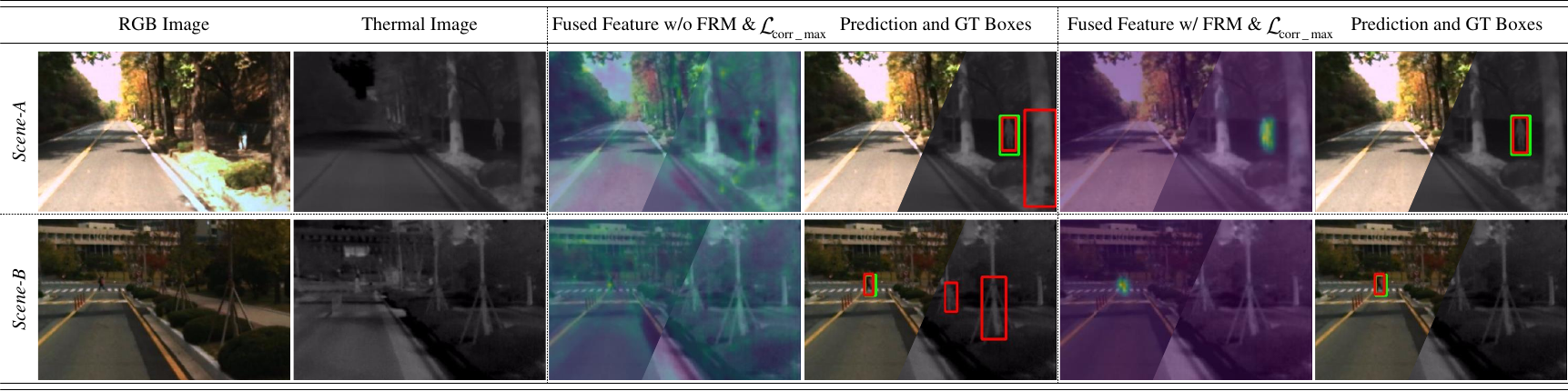}
		\caption{Comparison of feature visualizations and detection results between our method and a variant that doesn't use FRM and $\mathcal{L}_{\rm corr\_max}$.}
		\label{fig:frmFeature}
	\vspace{-12pt}
	\end{figure*}
	%%%%%%%%%%%%%%%%%%%%%%%%%%%%%%%%
	\begin{figure}[t]
		\centering
		\includegraphics[width=0.6\linewidth]{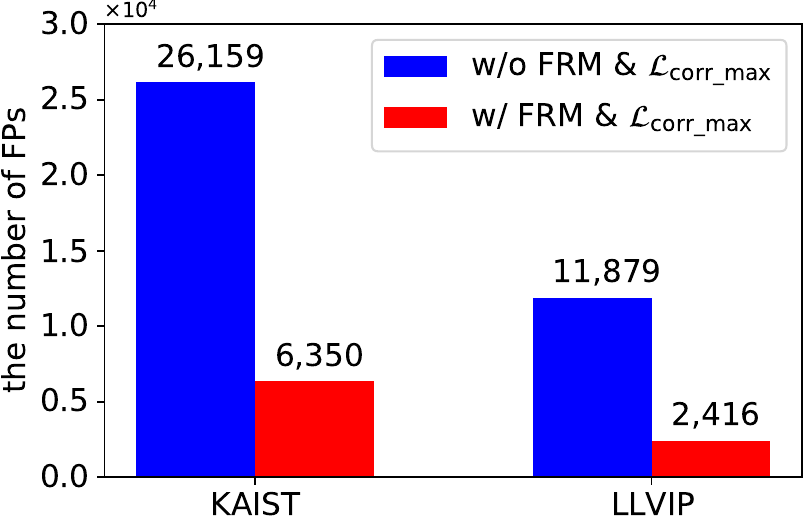}
		\caption{Comparison of the number of FPs between method that uses FRM and $\mathcal{L}_{\rm corr\_max}$ and the method that doesn't use them on two datasets.}
		\label{fig:frm_fps}
	\vspace{-12pt}
	\end{figure}
	%%%%%%%%%%%%%%%%%%%%%%%%%%%%%%%%
	
	Based on the analysis in Section~\ref{sec:model formulation}, our feature contrast enhancement method needs to ensure that: (1) the predicted box-level mask $\mathbf{m}$ is as accurate as possible, and (2) the processed correlation $\mathbf{s}$ between the predicted box-level mask and the initially fused feature is as high as possible. %Since the ground-truth box-level mask is binary and exhibits distinct contrast between the foreground and background, this method enables the detector to learn contrast-enhanced features.
	
	To this end, we propose a novel loss function named the correlation-maximum loss function $\mathcal{L}_{\rm corr\_max}$ to regularize the model. It is defined as
	\begin{equation}
		\mathcal{L}_{\rm corr\_max}(\mathbf{\tilde{m}}, \mathbf{m},\mathbf{s}) = \mathcal{L}_{\rm seg}(\mathbf{\tilde{m}}, \mathbf{m}) + \alpha \mathcal{L}_{\rm neg\_corr}(\mathbf{s}),
	\end{equation}
	where $\mathbf{\tilde{m}}$ denotes the GT box-level mask and $\alpha=\text{0.1}$ in this work.
	 This function comprises two components: the segmentation loss function $\mathcal{L}_{\rm seg}$ and the negative correlation loss function $\mathcal{L}_{\rm neg\_corr}$.
	 
	 The segmentation loss function $\mathcal{L}_{\rm seg}$ aims to maximize the accuracy of the predicted box-level mask. It is defined as
	 \begin{equation}
 		\mathcal{L}_{\rm seg}(\mathbf{\tilde{m}}, \mathbf{m})=\mathcal{L}_{\rm bce}(\mathbf{\tilde{m}}, \mathbf{m}) + \mathcal{L}_{\rm dice}(\mathbf{\tilde{m}}, \mathbf{m}).
	 \end{equation}
	The binary cross-entropy loss function $ \mathcal{L}_{\rm bce} $~\cite{focusnet} is defined as
 	\begin{equation}
 		\mathcal{L}_{\rm bce}(\mathbf{\tilde{m}}, \mathbf{m}) = \frac{1}{h\times w}\sum_{p=1}^{h\times w}l(\tilde{m}_p, m_p),
 	\end{equation}
 where 	 \begin{equation}
 	l(\tilde{m}_p, m_p)=-[\tilde{m}_p\log m_p\\+\left(1-\tilde{m}_p\right)\log\left(1-m_p\right)].
 \end{equation}
  	The dice loss function $ \mathcal{L}_{\rm dice} $~\cite{diceloss} is defined as
 	\begin{equation}
 		\begin{aligned}
 		\mathcal{L}_{\rm dice}(\mathbf{\tilde{m}}, \mathbf{m})&=1-\frac{2\left|\mathbf{\tilde{m}}\cap\mathbf{m}\right|+\epsilon}{\left|\mathbf{\tilde{m}}\right|+\left|\mathbf{m}\right|+\epsilon}\\
 		&=1-\frac{\left(2\times\sum_{p=1}^{h\times w}\tilde{m}_p m_p\right)+\epsilon}{\left(\sum_{p=1}^{h\times w}\tilde{m}_p\right)+\left(\sum_{p=1}^{h\times w}m_p\right)+\epsilon},
 		\end{aligned}
 	\end{equation}
 	where $m_p$ and $\tilde{m}_p$ denote the elements of $\mathbf{m}$ and $\mathbf{\tilde{m}}$ at the location $p$, respectively. $\epsilon$ is set to 1.0 following the setting in~\cite{diceloss}. We use nearest neighbor method to down-sample the GT box-level mask to match the shape of the predicted mask. $\mathbf{\tilde{m}}\cap\mathbf{m}$ computes their intersection, and $|\cdot|$ computes the magnitude of a mask.
	 
	 The negative correlation loss function $\mathcal{L}_{\rm neg\_corr}$ aims to make each entry in the processed correlation $\mathbf{s}$ approach 1.0. It is defined as
	\begin{equation}
	\mathcal{L}_{\rm neg\_corr}(\mathbf{s})=-\frac{1}{c}\sum_{i=1}^{c}\log s_i.
	\end{equation}
	Since the logarithm function often generates larger gradients when $s_i$ is close to zero compared to the $l_1$-norm-based (e.g., mean absolute error) and $l_2$-norm-based (e.g., mean squared error) loss functions, we construct the negative correlation loss function in the log domain. We combine the correlation-maximum losses $\mathcal{L}_{\rm corr\_max}$ from multiple CNN stages with the final detection losses, including classification and regression losses, to train the detectors.
	
	\textbf{Effect Validation: FRM and Correlation-Maximum Loss Function.} The purpose of our proposed FRM and the correlation-maximum loss function is to reduce FPs by enhancing the feature contrast between foreground and background regions. To understand their effects, we compare feature visualization results between the method with and without FRM and the correlation-maximum loss function, respectively, denoted as ``w/ FRM $\&\ \mathcal{L}_{\rm corr\_max}$'' and ``w/o FRM $\&\ \mathcal{L}_{\rm corr\_max}$'' for simplicity. Fig.~\ref{fig:frmFeature} shows that FRM and the correlation-maximum loss function effectively enhance the feature contrast while the counterpart generates noisy feature maps and FPs. Furthermore, we count the FPs across entire datasets. The results in Fig.~\ref{fig:frm_fps} show that our method significantly reduces the FPs on different datasets.
	
	%-------------------------------------------------
	\section{Experiments}
	\label{sec:experiments}
	In the experiments, we evaluate the performance of TFDet on four representative multispectral detection datasets, two of which are used for pedestrian detection while the other two datasets are used for multi-class object detection. We first describe these datasets, then introduce the implementation details on these datasets. Next, we extensively compare both the detection performance and inference time of TFDet with the previous state-of-the-art approaches. Lastly, we present the results of ablation studies.
	%-------------------------------------------------
%	% ==============Dataset Statistics======================
%	\begin{table}
%	\vspace{10pt}
%	\setlength{\tabcolsep}{8pt}
%	\renewcommand{\arraystretch}{1.2}
%	\caption{\color{blue}The statistics of the datasets used in this work.} 
%	\label{tbl:dataset}
%	\vspace{-12pt}
%	{\color{blue}
%	\begin{center}
%	\resizebox{\linewidth}{!}{
%	\begin{tabular}{l|c|c|c}
%	\hline\hline
%	Dataset & Categories & Train Images & Val Images\\\hline
%	KAIST~\cite{kaist} & Person	& 7,601 pairs	& 2,252 pairs\\\hline
%	LLVIP~\cite{llvip}	& Person	&12,025 pairs	& 3,463 pairs\\\hline
%	FLIR~\cite{Flir}	& {\begin{tabular}[c]{@{}c@{}}Person, Car, Bicycle	\end{tabular}}	& 4,129 pairs	& 1,013 pairs\\\hline
%	M3FD~\cite{TarDAL}	& {\begin{tabular}[c]{@{}c@{}}Person, Car, Bus,\\Motorcycle, Lamp, Truck\end{tabular}} & 2,905 pairs &1,295 pairs\\\hline\hline	
%	\end{tabular}}
%	\end{center}}
%	\vspace{-12pt}
%	\end{table}
%	% ======================================================
	% ================== KAIST============
	\begin{figure*}
	\begin{minipage}{\linewidth}
	\begin{center}
	\setlength{\tabcolsep}{16pt}
	\renewcommand{\arraystretch}{1.1}
	\captionof{table}{Comparison of MR (\%) on the KAIST dataset. We use Faster R-CNN~\cite{fasterrcnn} with VGG-16~\cite{vgg} as the baseline detector. The best result on each subset is highlighted in bold and marked in {\color[RGB]{255, 0, 0}\textbf{red}}, while the second-best result is underlined and marked in {\color[RGB]{0,128,0}\textbf{green}}. The six subsets can be categorized into two groups: (1) Scene - All-Day, Day, and Night; and (2) Distance - Near, Medium, and Far. The Scene subset results are evaluated on a reasonable set where pedestrians are not or partially occluded and have a height higher than 55 pixels. In the Distance subset, only non-occluded pedestrians are evaluated, with height categories of [115, +$\infty$), [45, 115), [1, 45) for Near, Medium, and Far, respectively.}
	\label{tbl:kaistComparison}
	\resizebox{\linewidth}{!}{
	\begin{tabular}{l|l|ccc|ccc}
	\hline\hline
	Method          &Publication Year                              & All-Day($\downarrow$) & Day($\downarrow$)     & Night($\downarrow$)   & Near($\downarrow$)   & Medium($\downarrow$)  & Far($\downarrow$)     \\ \hline
	\multicolumn{8}{c}{RGB-Based Detection Approaches}\\\hline
	{YOLOv5~\cite{yolov5}}         & {version 7.0}  & {22.68} & {16.15} & {36.26} & {2.37} & {30.04} & {68.27} \\\hline
	{Faster R-CNN~\cite{fasterrcnn}}     & {NeurIPS 2015} & {18.86} & {13.78} & {29.70} & {1.79} & {28.45} & {71.90}\\\hline
	{DeformableDETR~\cite{DeformableDETR}} & {ICLR 2021}    & {21.90} & {16.26} & {32.95} & {3.28} & {30.08} & {66.60} \\\hline
	{DINO~\cite{DINO}}           & {ICLR 2023}    & {25.99} & {20.14} & {38.34} & {3.65} & {31.35} & {65.62} \\\hline
	\multicolumn{8}{c}{{Thermal-Based Detection Approaches}}\\\hline
	{YOLOv5~\cite{yolov5}}         & {version 7.0}  & {15.75} & {21.52} & {5.62}  & {1.23} & {22.43} & {51.13} \\\hline
	{Faster R-CNN~\cite{fasterrcnn}}     & {NeurIPS 2015} & {13.93} & {18.79} & {4.93}  & {1.34} & {20.95} & {54.47} \\\hline
	{DeformableDETR~\cite{DeformableDETR}} & {ICLR 2021}    & {20.54} & {26.5}  & {9.44}  & {3.13} & {26.38} & {55.18} \\\hline
	{DINO~\cite{DINO}}           & {ICLR 2023}    & {23.24} & {29.19} & {11.73} & {8.03} & {25.72} & {51.67}\\\hline
	\multicolumn{8}{c}{Multispectral Detection Approaches}\\\hline
	IAF-RCNN~\cite{illuminationaware}& Pattern Recognition 2019             & 15.73 & 14.55 & 18.26 & 0.96 & 25.54 & 77.84 \\ \hline
	IATDNN+IAMSS~\cite{GUAN2019148} &Information Fusion 2019               & 14.95 & 14.67 & 15.72 & 0.04 & 28.55 & 83.42 \\ \hline
	IAHDANet~\cite{adaptation} &TITS 2023&14.65&18.88&5.92&-&-&-\\\hline
	CIAN~\cite{cian}& Information Fusion 2019                      & 14.12 & 14.77 & 11.13 & 3.71 & 19.04 & 55.82 \\ \hline
	MSR~\cite{msr}& AAAI 2022                               & 11.39 & 15.28 & 6.48  & -      & -       & -        \\ \hline
	TINet~\cite{TINet} &TIM 2023	&10.25 &7.48 &9.15 &- &- &- \\\hline
	ARCNN~\cite{arcnn} &ICCV 2019                            & 9.34  & 9.94  & 8.38  & \textbf{\color{red}0.00} & 16.08 & 69.00 \\ \hline
	{ARCNN-Extension~\cite{ARCNN1}} & {TNNLS 2021} & {9.03} & {9.79} & {8.07} & {-} & {-} & {-}\\\hline
	{C2Former~\cite{C2Former}} & {TGRS 2024} & {8.59} & {9.79} & {5.66} & \textbf{\color{red} 0.00} & {13.71} & {48.14}\\\hline
	{CMM~\cite{CMM}} & {CVPR 2024} & {8.54} & {9.6} & {5.93} & {-} & {-} & {-}\\\hline	
	CMPD~\cite{cmpd} &TMM 2022           & 8.16  & 8.77  & 7.31  & \textbf{\color{red}0.00} & 12.99 & 51.22 \\ \hline
	MBNet~\cite{mbnet}& ECCV 2020                          & 8.13  & 8.28  & 7.86  & \textbf{\color{red}0.00} & 16.07 & 55.99 \\ \hline
	SCDNet~\cite{scdnet}& TITS 2022 & 8.07  & 8.16  & 7.51  & -      & -       & -       \\ \hline
	{RITA~\cite{RITA}} & {TIV 2024} & {7.64} & {7.73} & {7.11} &{\color[RGB]{0, 128, 0}\underline{0.02}} & {10.47} & {40.9}\\\hline
	BAANet~\cite{baanet}& ICRA 2022                            & 7.92  & 8.37  & 6.98  & \textbf{\color{red}0.00} & 13.72 & 51.25 \\ \hline
	UCG~\cite{uffucg} &TCSVT 2022                          & 7.89  & 8.00  & 6.95  & 1.07 & 11.36 & {\color[RGB]{0, 128, 0}\underline{37.16}} \\ \hline
	MFPT~\cite{mfpt}& TITS 2023 & 7.72 & 8.26 & 4.53 & - & - & - \\ \hline
	MLPD~\cite{mlpd}& IEEE RA-L 2021                      & 7.58  & 7.95  & 6.95  & \textbf{\color{red}0.00} & 12.10 & 52.79 \\ \hline
	MSDS-RCNN~\cite{msds}& BMVC 2018                         & 7.49  & 8.09  & 5.92  & 1.07 & 12.33 & 58.55\\ \hline
	SMPD~\cite{SMPD}	&TCSVT 2023	&7.44	&8.12	&6.23	&\textbf{\color{red}0.00}	&11.04	&54.92\\\hline
	{CAT+MFT~\cite{CAT+MFT}} & {ECCV 2022} & {7.03} & {7.51} & {6.53} & {-} & {-} & {-}\\\hline
	{MCHE-CF~\cite{MCHE-CF}} & {TMM 2023} & {6.71} & {7.58} & {5.52} & \textbf{\color{red} 0.00} & {10.4} & \textbf{\color{red} 36.75}\\\hline
	{CPFM~\cite{CPFM}} & {TMM 2024} & {6.62} & {7.09} & {5.61} & {-} & {-} & {-}\\\hline
	GAFF~\cite{gaff} &WACV 2021                              & 6.48  & 8.35  & {\color[RGB]{0, 128, 0}\underline{3.46}}  & \textbf{\color{red}0.00} & 13.23 & 46.87  \\ \hline
	VTFYOLO~\cite{vtfyolo}&TII 2024&6.48&6.65&5.82&-&-&-\\ \hline
	{AANet~\cite{AANet}} & {ACM MM 2023} & {6.11} & {5.94} & {6.37} & {-} & {-} & {-}\\\hline
	{YOLO-Adaptor~\cite{YOLO-Adaptor}} & {TIV 2024} & {6.11} & {-} & {-} & {-} & {-} & {-}\\\hline
	DCMNet~\cite{dcmnet} &ACM MM 2022                  & 5.84  & 6.48  & 4.60  & {\color[RGB]{0, 128, 0}\underline{0.02}} & 16.07 & 69.70 \\ \hline
	{M2FNet~\cite{M2FNet}} & {TMM 2024} & {5.69} & {6.93} & {3.55} & \textbf{\color{red}0.00} & {\color[RGB]{0, 128, 0} \underline{8.55}} & {42.02}\\\hline
	ProbEn3 w/GAFF~\cite{proben}& ECCV 2022                    & 5.14  & 6.04  & 3.59  & \textbf{\color{red}0.00} & 9.59  & 41.92\\ \hline
	LG-FAPF~\cite{lgfapf}& Information Fusion 2022                    & {\color[RGB]{0, 128, 0}\underline{5.12}}  & {\color[RGB]{0, 128, 0}\underline{5.83}}  & 3.69  & 0.58 & \textbf{\color{red}8.44}  & 40.47 \\ \hline
	\rowcolor{gray!20}TFDet (Ours)  & -                                     & \textbf{\color{red}4.47}  & \textbf{\color{red}5.22}  & \textbf{\color{red}3.36}  & \textbf{\color{red}0.00} & 9.29  & 55.50\\\hline\hline
	\end{tabular}}
	\end{center}
	\end{minipage}
	\begin{minipage}{\linewidth}
	\centering
	\includegraphics[width=\linewidth]{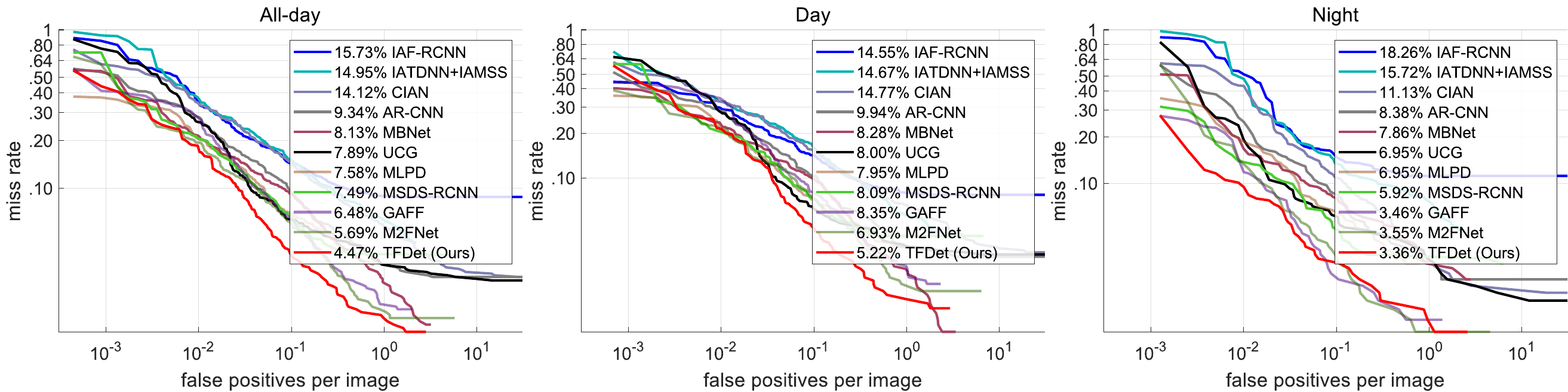}\\
	\caption{Comparison of miss rates with respect to false positives per image (FPPI) on the KAIST test set.}
	\label{fig:mrvsfppi}
	\end{minipage}
	\end{figure*}

	%-------------------------------------------------
	\begin{figure*}[t]
		\centering
		\includegraphics[width=0.9\linewidth]{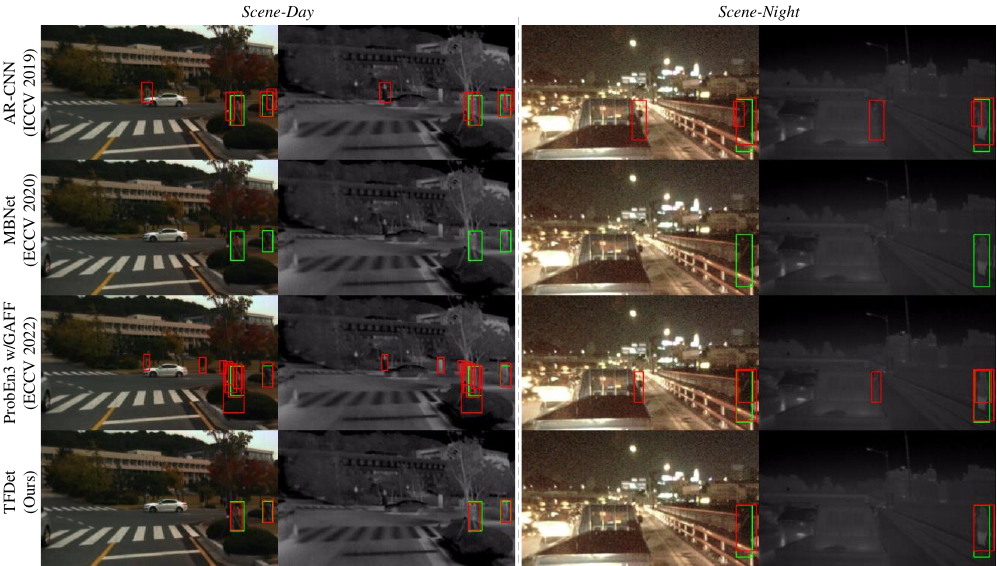}
		\caption{Qualitative comparison of TFDet with three state-of-the-art methods~\cite{arcnn, mbnet, proben} on two example scenes from the KAIST dataset. Ground-truth boxes are indicated with green bounding boxes, and predicted boxes are indicated with red bounding boxes.}
		\label{fig:qualitative}
		\vspace{-12pt}
	\end{figure*}
	%-------------------------------------------------
	\subsection{Datasets}
	\label{sec:dataset}
	
%	{\color{blue}
%	We present the important statistics of the datasets we used in Table~\ref{tbl:dataset}, and we provide more details about them below.}
	
	The \textbf{KAIST}~\cite{kaist} dataset is a widely used multispectral pedestrian detection dataset. Recent works~\cite{msds, halffusion} have updated this dataset because original version contains noisy annotations. The updated dataset includes 7,601 pairs of aligned RGB and thermal images in the training set and 2,252 in the validation set, each with a spatial resolution of 512~$\times$~640.
	
	The \textbf{LLVIP}~\cite{llvip} dataset is a more challenging multispectral pedestrian detection dataset. It is taken under low-light conditions, making it difficult to detect pedestrians in the RGB modality. This dataset includes 12,025 pairs of aligned RGB-T images in the training set and 3,463 in the validation set, with each image having a resolution of 1,024~$\times$~1,280.
	
	The \textbf{FLIR}~\cite{Flir} dataset is a benchmark in multispectral object detection. Recent work~\cite{AlignedFLIR} updates this dataset because the original version contains unaligned image pairs. We use the updated, aligned version following previous works~\cite{IGT, Fusion-Mamba} for a fair comparison. The aligned version consists of three categories: `person', `car', and `bicycle'. It has 4,129 pairs of RGB and thermal images in the training set and 1,013 pairs of RGB and thermal images in the validation set, with each image having a resolution of 512 $\times$ 640.
	
	The \textbf{M3FD}~\cite{TarDAL} dataset is a multispectral object detection benchmark. It consists of six categories: `person', `car', `bus', `motorcycle', `lamp', and `truck'. This dataset doesn't provide an official data split, thus we use the data split introduced in the recent work~\cite{EME}. The training and validation sets consist of 2,905 and 1,295 image pairs, respectively. Since the image resolutions in this dataset vary, we resize the longer side and pad the shorter side to obtain images with a resolution of 640 $\times$ 640.
	%-------------------------------------------------
	\subsection{Implementation Details}\label{sec:implementation}
	In our experiments, we employ Faster R-CNN~\cite{fasterrcnn} and YOLOv5~\cite{yolov5} as two-stage and one-stage detectors, respectively. We implement Faster R-CNN using the MMDetection~\cite{mmdetection} toolbox and YOLOv5 using its official repository~\cite{yolov5}. Unless otherwise specified, all experiments are run on two GTX 3090 GPUs. For Faster R-CNN, we train for 12 epochs with a total batch size of 6 and an initial learning rate of 0.01. The learning rate is decayed by a factor of 0.1 at epoch 8 and epoch 11. Unless otherwise specified, we use SGD as the optimizer and random horizontal flipping as the data augmentation technique. For YOLOv5, we use the hyperparameters specified in the official configuration file. In addition, to ensure a fair comparison with existing approaches, we adopt the same detectors, backbone networks, and evaluation metrics used in prior works.
	
	For the \textbf{KAIST} dataset, we adopt Faster R-CNN with VGG-16~\cite{vgg} as the baseline detector, following the practices in~\cite{illuminationaware, halffusion, GUAN2019148, cian, msds, arcnn, dcmnet, lgfapf}. To evaluate the performance of detectors, we calculate the log-average miss rate (MR)~\cite{piotr} using a commonly-used toolbox provided in~\cite{msds}. A lower MR indicates better detection performance.
	
	For the \textbf{LLVIP} dataset, we use Faster R-CNN with ResNet-50~\cite{resnet} and YOLOv5-Large~\cite{yolov5} as the baseline detectors, following the practices in~\cite{detfusion, dcmnet, llvip}. We note that YOLOv5 is trained for 100 epochs. Model performance is evaluated using the average precision metric, with AP50 denoting the average precision at $ \rm IoU=0.50 $. The AP is calculated by averaging over 10 IoU thresholds ranging from 0.50 to 0.95 with 0.05 interval. A higher AP indicates better detection performance.
	
	For the \textbf{FLIR} dataset, we use Faster R-CNN~\cite{fasterrcnn} as the baseline detector and AP as the evaluation metric. We use Swin-Transformer-Tiny~\cite{swin} as the backbone network and AdamW as the optimizer. We set the initial learning rate to $\text{10}^{\text{-4}}$, and use random horizontal flipping and random cropping as data augmentation techniques, following the implementations in previous state-of-the-art work~\cite{IGT}.
	
	For the \textbf{M3FD} dataset, we use YOLOv5-Small~\cite{yolov5} as the baseline detector and AP as the evaluation metric. We train YOLOv5-Small for 36 epochs. We re-implement all comparative approaches using the same data split for a fair comparison. For image fusion-based approaches, we first generate fused images based on their codes, then train YOLOv5-Small using the same settings as our method.
	%-------------------------------------------------
	\subsection{Comparison on the KAIST Dataset}
	We conduct experiments on the KAIST dataset and compare our TFDet with previous state-of-the-art approaches from both quantitative and qualitative perspectives.
	
	In Table~\ref{tbl:kaistComparison}, we present the quantitative results of comparative approaches and our TFDet, where both single-modality and multispectral detection approaches are included. The results demonstrate that our TFDet achieves the best detection performance. Specifically, TFDet obtains a 9.46\% absolute performance gain compared to the best single-modality approach, Faster R-CNN~\cite{fasterrcnn}, and a 0.65\% gain over the previous state-of-the-art multispectral approach, LG-FAPF~\cite{lgfapf}. The results of M2FNet~\cite{M2FNet} are higher than those reported in the original paper because the original results are obtained by training for 300 epochs, while we train it for 12 epochs for a fair comparison. To the best of our knowledge, TFDet is the first method to achieve an MR value lower than 5\% with 12 epochs training. Notably, TFDet exhibits significant superiority in the Night subset, where RGB images are generally of poor visual quality.
	
	To comprehensively analyze these advantages, we plot the miss rate versus false positives per image (FPPI) curves in Fig.~\ref{fig:mrvsfppi}. The results show that TFDet misses fewer pedestrians even when the confidence threshold is high (i.e., FPPI is low). 
	
	We also conduct qualitative comparisons with previous state-of-the-art approaches. AR-CNN~\cite{arcnn}, MBNet~\cite{mbnet}, and ProbEn~\cite{proben} are selected as counterparts, whose results are publicly available. Fig.~\ref{fig:qualitative} presents these comparisons on two scenes (day and night). The results show that the counterparts generate unwanted FPs, such as red boxes in the background or replicated boxes on pedestrians, as well as false negatives (\emph{i.e.,} missed pedestrians). In contrast, TFDet successfully detects all pedestrians in both scenes.

	\subsection{Comparison on the LLVIP Dataset}
	%-------------------------------------------------
	\begin{table}
	\setlength{\tabcolsep}{1.2pt}
	\renewcommand{\arraystretch}{1.2}
	\caption{Comparison of AP (\%) on the LLVIP dataset. {The best results are highlighted in bold and marked in {\color[RGB]{255, 0, 0}\textbf{red}}, while the second-best results are underlined and marked in {\color[RGB]{0,128,0}\textbf{green}}.}}
	\label{tbl:llvip}
	\vspace{-12pt}
	\begin{center}
	\resizebox{\linewidth}{!}{
	\begin{tabular}{l|l|cc}
	\hline\hline
	Method         & Publication Year & $\text{AP}(\uparrow)$ & AP50$(\uparrow)$\\\hline
	\multicolumn{4}{c}{RGB-Based Detection Approaches}\\\hline
	{IEGOD~\cite{IEGOD}}          & {TNNLS 2023}   & {-}    & {87.6}\\\hline
	{DeformableDETR~\cite{DeformableDETR}} & {ICLR 2021}    & {45.5} & {88.7}\\\hline
	{FasterRCNN~\cite{fasterrcnn}}     & {NeurIPS 2015} & {49.2} & {90.1}\\\hline
	{DINO~\cite{DINO}}           & {ICLR 2023}    & {52.3} & {90.5} \\\hline
	YOLOv5~\cite{yolov5}         & version 7.0  & 52.7 & 90.8 \\\hline
	\multicolumn{4}{c}{Thermal-Based Detection Approaches}\\\hline
	{DINO~\cite{DINO}}           & {ICLR 2023}    & {51.3} & {90.2} \\\hline
	{HalluciDet~\cite{HalluciDet}}     & {WACV 2024}    & {57.8} & {90.1} \\\hline
	{FasterRCNN~\cite{fasterrcnn}}     & {NeurIPS 2015} & {58.3} & {94.3} \\\hline
	{DeformableDETR~\cite{DeformableDETR}} & {ICLR 2021}    & {61.9} & {96.1} \\\hline
	{TIRDet~\cite{TIRDet}}         & {ACM MM 2023}  & {64.2} & {96.3}\\\hline
	YOLOv5~\cite{yolov5}              & version 7.0              & {\color[RGB]{0, 128, 0} \underline{67.0}}    & {96.5}  \\\hline
	\multicolumn{4}{c}{Multispectral Detection Approaches}\\\hline
	{PoolFuser~\cite{PoolFuser}}              & {AAAI 2023}                & {38.4}          & {80.3}          \\\hline
	DetFusion~\cite{detfusion}              & ACM MM 2022              & -             &80.7          \\\hline
	ProbEn~\cite{proben}                 & ECCV 2022                & 51.5          & 93.4          \\\hline
	{DIVFusion~\cite{DIVFusion}}              & {Information Fusion 2023}  & {52.0}            & {89.8}          \\\hline
	%	{\color{blue} DDcGAN~\cite{DDcGAN}}                 & {\color{blue} TIP 2020}                 & {\color{blue} 52.0}            & {\color{blue} 89.8}          \\\hline
	{LENFusion~\cite{LENFusion}}              & {TIM 2024}                 & {53.0}            & {81.6}          \\\hline
	{DM-Fusion~\cite{DM-Fusion}}              & {TNNLS 2023}               & {53.1}          & {88.1}          \\\hline
	{CAMF~\cite{CAMF}}                   & {TMM 2024}                 & {55.6}          & {89.0}            \\\hline
	{ARCNN-Extension~\cite{ARCNN1}}       & {TNNLS 2021}               & {56.23}         & {-}          \\\hline
	{MoE-Fusion~\cite{MoE-Fusion}}             & {ICCV 2023}                & {-}             & {91.0}            \\\hline
	%	{\color{blue} TarDAL~\cite{TarDAL}}                 & {\color{blue} CVPR 2022}                & {\color{blue} 56.4}          & {\color{blue} 91.2}         \\\hline
	{MetaFusion~\cite{MetaFusion}}             & {CVPR 2023}                & {56.9}          & {91.0}            \\\hline
	%	{\color{blue} U2Fusion~\cite{U2Fusion}}               & {\color{blue} TPAMI 2020}               & {\color{blue} 57.6}          & {\color{blue} 90.7}         \\\hline
	{TFNet~\cite{TFNet}}                  & {TITS 2023}                & {57.6}          & {-}           \\\hline
	{DDFM~\cite{DDFM}}                  & {ICCV 2023}                & {58.0}            & {91.5}   \\\hline
	DCMNet~\cite{dcmnet}                 & ACM MM 2022              & 58.4          & -           \\\hline
	{CSAA~\cite{CSAA}} & {CVPR 2023} & {59.2} & {94.3} \\\hline
	{Diff-IF~\cite{Diff-IF}}                & {Information Fusion 2024}  & {59.5}          & {93.3}         \\\hline
	{YOLO-Adaptor~\cite{YOLO-Adaptor}}           & {TIV 2024}                 & {-}             & {96.5}      \\\hline
	{Fusion-Mamba~\cite{Fusion-Mamba}}           & {arXiv 2024}               & {62.8}          & {\color[RGB]{0, 128, 0} \underline{96.8}}     \\\hline
	{CALNet~\cite{CALNet}}                 & {ACM MM 2023}              & {63.4}          & {-}          \\\hline
	{LRAF-Net~\cite{LRAF-Net}}               & {TNNLS 2023}               & {66.3}          & {\color{red} \textbf{97.9}}  \\\hline
	\rowcolor{gray!20}{TFDet-FasterRCNN (Ours)} & {-}                        & {59.4}          & {96.0}       \\\hline
	\rowcolor{gray!20}{TFDet-YOLOv5 (Ours)}    & {-}                        & {\color{red} \textbf{71.1}} & {\color{red} \textbf{97.9}}\\\hline\hline
	\end{tabular}}
	\end{center}
	\vspace{-12pt}
	\end{table}

	In Table~\ref{tbl:llvip}, we evaluate TFDet on the LLVIP dataset and compare it with previous state-of-the-art approaches, including both single-modality and multispectral detection approaches. We employ both Faster R-CNN and YOLOv5 as the baseline detectors to construct TFDet. The results demonstrate that TFDet using YOLOv5 (TFDet-YOLOv5) achieves the best performance. Specifically, TFDet-YOLOv5 outperforms the best single-modality approach by 4.1\% AP, and outperforms the best multispectral approach, LRAF-Net~\cite{LRAF-Net}, by 4.8\% AP. Interestingly, YOLOv5 using thermal images exhibits good performance among previous approaches. This is because most of the images in the LLVIP dataset are captured under low-light scenes, where thermal images are sufficient to detect most pedestrians in this dataset. Nevertheless, our TFDet-YOLOv5 exceeds previous approaches by a large margin, benefiting from the significant reduction of FPs, as shown in Fig.~\ref{fig:frm_fps}.
	%-----------------
	%======================FLIR Dataset=====================
	\begin{table}[t]
	\setlength{\tabcolsep}{1pt}
	\renewcommand{\arraystretch}{1.2}
	\caption{Comparison of AP (\%) on the FLIR dataset. The best results are highlighted in bold and marked in {\color[RGB]{255, 0, 0}\textbf{red}}, while the second-best results are underlined and marked in {\color[RGB]{0,128,0}\textbf{green}}.}
	\label{tbl:FLIR}
	\vspace{-12pt}
	\begin{center}
	\resizebox{\linewidth}{!}{
	\begin{tabular}{l|l|cc}
	\hline\hline
	Method         & Publication Year & $\text{AP}(\uparrow)$ & AP50$(\uparrow)$\\\hline
	\multicolumn{4}{c}{RGB-Based Detection Approaches}\\\hline
	{FasterRCNN~\cite{fasterrcnn}}     & {NeurIPS 2015} & {30.2} & {67.6}\\\hline
	{DINO~\cite{DINO}}           & {ICLR 2023}    & {30.5} & {65.3} \\\hline
	{DeformableDETR~\cite{DeformableDETR}} & {ICLR 2021}    & {31.2} & {68.4}\\\hline
	{YOLOv5~\cite{yolov5}}         & {version 7.0}  & {32.3} & {67.9}\\\hline
	\multicolumn{4}{c}{Thermal-Based Detection Approaches}\\\hline
	{EGMT~\cite{EGMT}}           & {ICRA 2023}    & {23.9} & {40.8}\\\hline
	{CE-RetinaNet~\cite{CE-RetinaNet}}   & {TGRS 2023}    & {30.0} & {62.3}\\\hline
	{DeformableDETR~\cite{DeformableDETR}} & {ICLR 2021}    & {39.8} & {76.6}\\\hline
	{DINO~\cite{DINO}}           & {ICLR 2023}    & {40.3} & {78.7}\\\hline
	{FasterRCNN~\cite{fasterrcnn}}     & {NeurIPS 2015} & {40.6} & {78.5} \\\hline
	{YOLOv5~\cite{yolov5}}         & {version 7.0}  & {42.2} & {78.4}\\\hline
	{TIRDet~\cite{TIRDet}}         & {ACM MM 2023}  & {44.3} & {81.4}\\\hline
	\multicolumn{4}{c}{Multispectral Detection Approaches}\\\hline
	{MoE-Fusion~\cite{MoE-Fusion}}     & {ICCV 2023}                    & {-}    & {55.8} \\\hline
	%		{GAFF~\cite{gaff}}           & {WACV 2021}                    & {37.4} & {74.6} \\\hline
	{ProbEn~\cite{proben}}         & {ECCV 2022}                    & {37.9} & {75.5}  \\\hline
	{MSAT~\cite{MSAT}}           & {IEEE SPL 2023}                & {39.0}   & {76.2} \\\hline
	{CSAA~\cite{CSAA}}           & {CVPR 2023}                    & {41.3} & {79.2} \\\hline
	{ICAFusion~\cite{ICAFusion}}      & {Pattern Recognition 2024}     & {41.4} & {79.2}  \\\hline
	{MFPT~\cite{mfpt}}           & {TITS 2023}                    & {-}    & {80.0}    \\\hline
	{YOLO-Adaptor~\cite{YOLO-Adaptor}}   & {TIV 2024}                     & {-}    & {80.1}  \\\hline
	{CMX~\cite{cmx}}            & {TITS 2023}                    & {42.3} & {82.2}\\\hline
	{LRAF-Net~\cite{LRAF-Net}}       & {TNNLS 2023}                   & {42.8} & {80.5}  \\\hline
	{IGT~\cite{IGT}}            & {Knowledge-Based Systems 2023} & {43.6} & {\color[RGB]{0, 128, 0}\underline{85.0}} \\\hline
	{Fusion-Mamba~\cite{Fusion-Mamba}}   & {arXiv 2024}                   & {\color[RGB]{0, 128, 0}\underline{44.4}} & {84.3} \\\hline
	\rowcolor{gray!20}{TFDet (Ours)}   & {-}                            & \textbf{\color{red}46.6} & \textbf{\color{red}86.6}\\\hline\hline
	\end{tabular}}
	\end{center}
	\vspace{-12pt}
	\end{table}
	% ================================================
	% ==========================M3FD dataset===========
	\begin{table}[t]
	\setlength{\tabcolsep}{5pt}
	\renewcommand{\arraystretch}{1.2}
	\caption{Comparison of AP (\%) on the M3FD dataset. All results are obtained using YOLOv5~\cite{yolov5} as the detector. The best results are highlighted in bold and marked in {\color[RGB]{255, 0, 0}\textbf{red}}, while the second-best results are underlined and marked in {\color[RGB]{0,128,0}\textbf{green}}.}
	\label{tbl:M3FD}
	\vspace{-12pt}
	\begin{center}
	\resizebox{\linewidth}{!}{
	\begin{tabular}{l|l|cc}
	\hline\hline
	Method         & Publication Year & $\text{AP}(\uparrow)$ & AP50$(\uparrow)$ \\\hline
	\multicolumn{4}{c}{Single-Modality Detection Baselines}\\\hline
	RGB~\cite{yolov5}	& version 7.0	& 36.1 & 60.2\\\hline
	Thermal~\cite{yolov5}	& version 7.0	& 34.9	&57.2 \\\hline
	\multicolumn{4}{c}{Multispectral Detection Approaches}\\\hline
	DIVFusion~\cite{DIVFusion}	& Information Fusion 2023 & 37.1	& 60.8 \\\hline
	PSFusion~\cite{PSFusion}	& Information Fusion 2023 &  38.0	& 61.1 \\\hline
	AUIF~\cite{AUIF}	& TCSVT 2022	& 38.3	&  {\color[RGB]{0, 128, 0}\underline{62.0}} \\\hline
	CDDF~\cite{CDDFuse}	& CVPR 2023	& 38.6	& 61.9\\\hline
	%	DDcGAN~\cite{DDcGAN} & TIP 2020	& 37.1	& 61.0 \\\hline
	%	DenseFuse~\cite{DenseFuse}	& TIP 2019	& 38.9	& {\color[RGB]{0, 128, 0}\underline{62.4}} \\\hline
	U2Fusion~\cite{U2Fusion}	& TPAMI 2020 & 38.7 	& 61.9 \\\hline
	TarDAL~\cite{TarDAL}	& CVPR 2022	& {\color[RGB]{0, 128, 0}\underline{39.1}} &61.9\\\hline
	\rowcolor{gray!20}TFDet (Ours)	& - 	& {\color{red}\textbf{41.0}} 	& {\color{red}\textbf{64.8}}  \\\hline\hline
	\end{tabular}}
	\end{center}
	\vspace{-12pt}
	\end{table}
	% ==================
	\subsection{Extension to Multi-Class Object Detection Scenarios}
	\label{sec:extension}
	We conduct experiments on two multispectral object detection datasets, FLIR~\cite{Flir} and M3FD~\cite{TarDAL}, to evaluate the effect of TFDet for multi-class object detection. We also compare TFDet with single-modality and multispectral detection approaches. Table~\ref{tbl:FLIR} and Table~\ref{tbl:M3FD} present experimental results on the FLIR and M3FD datasets, respectively.
	
	From Table~\ref{tbl:FLIR}, we have the following observations: (1) Our TFDet demonstrates a significant improvement (+2.3\% AP) over the best single-modality detection approach, TIRDet~\cite{TIRDet}. (2) Our TFDet significantly exceeds the previous state-of-the-art multispectral detection approach, Fusion-Mamba~\cite{Fusion-Mamba}, by 2.2\% AP. Table~\ref{tbl:M3FD} presents results on the M3FD dataset. Our TFDet also achieves state-of-the-art performance, outperforming the second-best approach, TarDAL~\cite{TarDAL}, by 1.9\% AP. These improvements verify that our target-aware fusion strategy benefits the multi-class object detection task.

	%-------------------------------------------------
	\subsection{Inference Time}
	
	\begin{table}
	\setlength{\tabcolsep}{10pt}
	\renewcommand{\arraystretch}{1.1}
	\caption{Inference time of Faster R-CNN based methods on the KAIST dataset at a resolution of 512~$\times$~640.}
	\vspace{-12pt}
	\begin{center}
	\resizebox{\linewidth}{!}{
	\begin{tabular}{l|c}
	\hline\hline
	Method&Time (seconds)\\ \hline
	MSDS-RCNN~\cite{msds} (BMVC 2018)$^\dag$	&0.22\\\hline
	IATDNN+IAMSS~\cite{GUAN2019148} (Information Fusion 2019)$^\dag$& 0.25 \\\hline
	IAF-RCNN~\cite{illuminationaware} (Pattern Recognition 2019)$^\dag$ &0.21\\\hline
	AR-CNN~\cite{arcnn} (ICCV 2019)$^\dag$	&0.17\\\hline
	DCMNet~\cite{dcmnet} (ACM MM 2022)$^\dag$	& 0.14 \\\hline
	\rowcolor{gray!20}TFDet (Ours)	&\textbf{0.13}\\\hline\hline
	\multicolumn{2}{l}{\rule{0in}{1.2em}$^\dag$\scriptsize Evaluation on a single TitanX GPU~\cite{dcmnet}.}
	\end{tabular}}
	\end{center}
	\label{tbl:speed}
	\vspace{-12pt}
	\end{table}
	
	We test the inference time of TFDet on the KAIST dataset at a resolution of 512~$\times$~640. Since we use Faster R-CNN~\cite{fasterrcnn} with VGG-16~\cite{vgg} as the baseline detector for this dataset, we compare our results with approaches that have the same settings~\cite{msds, GUAN2019148, illuminationaware, arcnn, dcmnet}. These approaches are tested on a single TitanX GPU according to~\cite{dcmnet}. To ensure a fair comparison, we use an RTX 3060 GPU in this experiment. According to the public documentation\footnote{\url{https://www.mydrivers.com/zhuanti/tianti/gpu/index.html}}, it is confirmed that they have equivalent computing power.

	Table~\ref{tbl:speed} presents the inference time of various methods. The results show that our TFDet has a comparable speed to the state-of-the-art counterpart DCMNet~\cite{dcmnet}.

	%-------------------------------------------------
	\subsection{Ablation Study}
	\label{sec:ablationstudy}
	\begin{figure}
	\begin{minipage}{\linewidth}
	%----tbl:ablationStructure
	\begin{center}
	\setlength{\tabcolsep}{10pt}
	\renewcommand{\arraystretch}{1.1}
	\captionof{table}{Ablation study results on the KAIST dataset for different components.}
	\label{tbl:ablationStructure}
	\vspace{-6pt}
	\resizebox{\linewidth}{!}{
	\begin{tabular}{lccccc}
	\hline\hline
	Method & FFM &FRM&$\mathcal{L}_{\rm seg}$&$\mathcal{L}_{\rm neg\_corr}$ & MR ($\downarrow$) \\\hline
	\multirow{3}{*}{Ablation}
	{}&\checkmark&&&&8.57\\\cline{2-6}
	{}&\checkmark&\checkmark&&&6.82\\\cline{2-6}
	{}&\checkmark&\checkmark&\checkmark&&5.41 \\\hline
	\rowcolor{gray!20}TFDet (Ours)&\checkmark&\checkmark&\checkmark&\checkmark&\textbf{4.47}\\
	\hline\hline
	\end{tabular}}
	\end{center}
	\vspace{0pt}
	\end{minipage}
	\begin{minipage}{\linewidth}
	%%%%%%%%%%%%%%%%%%%%%%%%%%%%%%%%
	\centering
	\includegraphics[width=\linewidth]{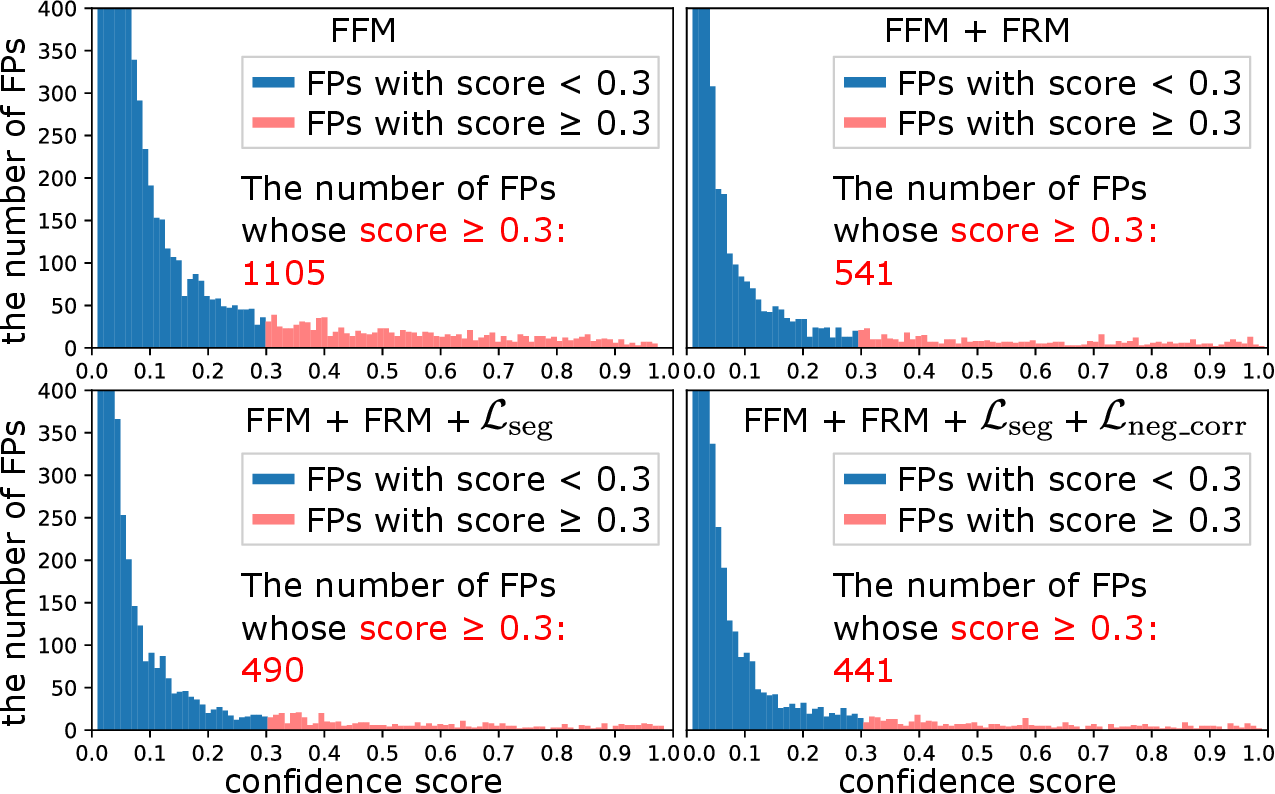}\\
	\caption{The confidence score histogram of false positives (FPs) on the KAIST dataset under various ablation study settings.}
	\label{fig:ablationfptp}
	\vspace{-12pt}
	\end{minipage}
	\end{figure}
	
	We conduct ablation studies on the KAIST dataset to evaluate the effect of each component in our method. Specifically, we progressively include each component (FFM, FRM, $\mathcal{L}_{\rm seg}$, and $\mathcal{L}_{\rm neg\_corr}$) during the fusion process and report the corresponding detection result in Table~\ref{tbl:ablationStructure}. Meanwhile, we show the confidence score histogram of FPs under each ablation study setting in Fig.~\ref{fig:ablationfptp}. In this figure, we mark the bars with confidence scores greater than 0.3 in red and display their quantities for easier comparison. The reason for choosing 0.3 is that FPs with scores greater than this value have a significantly more negative impact on detection performance. We can observe this phenomenon in Fig.~\ref{fig:performance vs fps}, where the log-average miss rate decrease rapidly when FPs with scores greater than $\text{10}^\text{-0.4}$($\approx$0.398) are removed.
	
	From Table~\ref{tbl:ablationStructure} and Fig.~\ref{fig:ablationfptp}, we have the following observations: (1) when including both the FFM and FRM, MR decreases by 1.75\% compared to using only FFM, and the number of FPs with scores greater than 0.3 decreases by 564, (2) when including FFM, FRM, and $\mathcal{L}_{\rm seg}$, MR decreases by 3.16\% compared to using only FFM, and the number of FPs with scores greater than 0.3 decreases by 615, and (3) when including all components, MR decreases by 4.1\% and the number of FPs decreases by 664 compared to using only FFM. These results confirm the relevance between the decrease in FPs and the improvement in detection performance.
	%-------------------------------------------------
	\section{Conclusions}\label{sec:conclusions}
	In this work, we address the challenge of feature fusion in the multispectral pedestrian detection task. We find that false positives (FPs) caused by noisy features significantly deteriorate detection performance, which has been overlooked by previous works. To address the issue of noisy features, we propose a target-aware fusion strategy for the detection task, named TFDet. Within this strategy, we introduce a feature fusion module (FFM) to collect complementary features, a feature refinement module (FRM) to distinguish between pedestrian and background features, and a correlation-maximum loss function to enhance feature contrast. Visualization results demonstrate that our fusion strategy generates discriminative features and significantly reduces FPs. As a result, TFDet achieves state-of-the-art performance on four benchmarks. Additionally, our strategy is flexible and computationally efficient, allowing it to be applied to both one-stage and two-stage detectors while maintaining low inference time. Based on the target-aware fusion strategy, we will explore multispectral object tracking in the future.
\bibliographystyle{IEEEtran}
\bibliography{zxbib.bib}

\begin{thebibliography}{10}
\providecommand{\url}[1]{#1}
\csname url@samestyle\endcsname
\providecommand{\newblock}{\relax}
\providecommand{\bibinfo}[2]{#2}
\providecommand{\BIBentrySTDinterwordspacing}{\spaceskip=0pt\relax}
\providecommand{\BIBentryALTinterwordstretchfactor}{4}
\providecommand{\BIBentryALTinterwordspacing}{\spaceskip=\fontdimen2\font plus
\BIBentryALTinterwordstretchfactor\fontdimen3\font minus
  \fontdimen4\font\relax}
\providecommand{\BIBforeignlanguage}[2]{{%
\expandafter\ifx\csname l@#1\endcsname\relax
\typeout{** WARNING: IEEEtran.bst: No hyphenation pattern has been}%
\typeout{** loaded for the language `#1'. Using the pattern for}%
\typeout{** the default language instead.}%
\else
\language=\csname l@#1\endcsname
\fi
#2}}
\providecommand{\BIBdecl}{\relax}
\BIBdecl

\bibitem{kaist}
S.~Hwang, J.~Park, N.~Kim, Y.~Choi, and I.~So~Kweon, ``{Multispectral
  Pedestrian Detection: Benchmark Dataset and Baseline},'' in \emph{Proceedings
  of the IEEE Conference on Computer Vision and Pattern Recognition (CVPR)},
  2015.

\bibitem{llvip}
X.~Jia, C.~Zhu, M.~Li, W.~Tang, and W.~Zhou, ``{LLVIP: A Visible-Infrared
  Paired Dataset for Low-Light Vision},'' in \emph{Proceedings of the IEEE
  International Conference on Computer Vision (ICCV)}, 2021.

\bibitem{lgfapf}
Y.~Cao, X.~Luo, J.~Yang, Y.~Cao, and M.~Y. Yang, ``{Locality Guided Cross-Modal
  Feature Aggregation and Pixel-Level Fusion for Multispectral Pedestrian
  Detection},'' \emph{Information Fusion}, 2022.

\bibitem{msds}
C.~Li, D.~Song, R.~Tong, and M.~Tang, ``{Multispectral Pedestrian Detection via
  Simultaneous Detection and Segmentation},'' in \emph{British Machine Vision
  Conference (BMVC)}, 2018.

\bibitem{halffusion}
J.~Liu, S.~Zhang, S.~Wang, and D.~Metaxas, ``{Multispectral Deep Neural
  Networks for Pedestrian Detection},'' in \emph{British Machine Vision
  Conference (BMVC)}, 2016.

\bibitem{dcmnet}
J.~Xie, R.~M. Anwer, H.~Cholakkal, J.~Nie, J.~Cao, J.~Laaksonen, and F.~S.
  Khan, ``{Learning a Dynamic Cross-Modal Network for Multispectral Pedestrian
  Detection},'' in \emph{Proceedings of the ACM International Conference on
  Multimedia (ACM MM)}, 2022.

\bibitem{baanet}
X.~Yang, Y.~Qian, H.~Zhu, C.~Wang, and M.~Yang, ``{BAANet: Learning
  Bi-Directional Adaptive Attention Gates for Multispectral Pedestrian
  Detection},'' in \emph{International Conference on Robotics and Automation
  (ICRA)}, 2022.

\bibitem{arcnn}
L.~Zhang, X.~Zhu, X.~Chen, X.~Yang, Z.~Lei, and Z.~Liu, ``{Weakly Aligned
  Cross-Modal Learning for Multispectral Pedestrian Detection},'' in
  \emph{Proceedings of the IEEE International Conference on Computer Vision
  (ICCV)}, 2019.

\bibitem{mbnet}
K.~Zhou, L.~Chen, and X.~Cao, ``{Improving Multispectral Pedestrian Detection
  by Addressing Modality Imbalance Problems},'' in \emph{Proceedings of the
  European Conference on Computer Vision (ECCV)}, 2020.

\bibitem{LRAF-Net}
H.~Fu, S.~Wang, P.~Duan, C.~Xiao, R.~Dian, S.~Li, and Z.~Li, ``{LRAF-Net:
  Long-Range Attention Fusion Network for Visible-Infrared Object Detection},''
  \emph{IEEE Transactions on Neural Networks and Learning Systems}, pp. 1--14,
  2023.

\bibitem{cmx}
J.~Zhang, H.~Liu, K.~Yang, X.~Hu, R.~Liu, and R.~Stiefelhagen, ``{CMX:
  Cross-Modal Fusion for RGB-X Semantic Segmentation with Transformers},''
  \emph{IEEE Transactions on Intelligent Transportation Systems}, 2023.

\bibitem{uffucg}
J.~U. Kim, S.~Park, and Y.~M. Ro, ``{Uncertainty-Guided Cross-Modal Learning
  for Robust Multispectral Pedestrian Detection},'' \emph{IEEE Transactions on
  Circuits and Systems for Video Technology}, 2022.

\bibitem{mfpt}
Y.~Zhu, X.~Sun, M.~Wang, and H.~Huang, ``{Multi-Modal Feature Pyramid
  Transformer for RGB-Infrared Object Detection},'' \emph{IEEE Transactions on
  Intelligent Transportation Systems}, 2023.

\bibitem{scdnet}
K.~Dasgupta, A.~Das, S.~Das, U.~Bhattacharya, and S.~Yogamani,
  ``{Spatio-Contextual Deep Network-Based Multimodal Pedestrian Detection for
  Autonomous Driving},'' \emph{IEEE Transactions on Intelligent Transportation
  Systems}, 2022.

\bibitem{proben}
Y.-T. Chen, J.~Shi, Z.~Ye, C.~Mertz, D.~Ramanan, and S.~Kong, ``{Multimodal
  Object Detection via Probabilistic Ensembling},'' in \emph{Proceedings of
  the European Conference on Computer Vision (ECCV)}, 2022.

\bibitem{Flir}
``{FREE FLIR Thermal Dataset for Algorithm Training},''
  \url{https://www.flir.com/oem/adas/adas-dataset-form/}.

\bibitem{TarDAL}
J.~Liu, X.~Fan, Z.~Huang, G.~Wu, R.~Liu, W.~Zhong, and Z.~Luo, ``{Target-Aware
  Dual Adversarial Learning and A Multi-Scenario Multi-Modality Benchmark to
  Fuse Infrared and Visible for Object Detection},'' in \emph{Proceedings of
  the Conference on Computer Vision and Pattern Recognition (CVPR)}, 2022, pp.
  5802--5811.

\bibitem{yolo}
J.~Redmon, S.~Divvala, R.~Girshick, and A.~Farhadi, ``{You Only Look Once:
  Unified, Real-Time Object Detection},'' in \emph{Proceedings of the IEEE
  Conference on Computer Vision and Pattern Recognition (CVPR)}, 2016.

\bibitem{detr}
N.~Carion, F.~Massa, G.~Synnaeve, N.~Usunier, A.~Kirillov, and S.~Zagoruyko,
  ``{End-to-End Object Detection with Transformers},'' in \emph{Proceedings of
  the European Conference on Computer Vision (ECCV)}, 2020.

\bibitem{fasterrcnn}
S.~Ren, K.~He, R.~Girshick, and J.~Sun, ``{Faster R-CNN: Towards Real-Time
  Object Detection with Region Proposal Networks},'' in \emph{Proceedings of
  the Advances in Neural Information Processing Systems (NeurIPS)}, 2015.

\bibitem{sparsercnn}
P.~Sun, R.~Zhang, Y.~Jiang, T.~Kong, C.~Xu, W.~Zhan, M.~Tomizuka, L.~Li,
  Z.~Yuan, C.~Wang, and P.~Luo, ``{Sparse R-CNN: End-to-End Object Detection
  with Learnable Proposals},'' in \emph{Proceedings of the IEEE Conference on
  Computer Vision and Pattern Recognition (CVPR)}, 2021.

\bibitem{detection20years}
Z.~Zou, K.~Chen, Z.~Shi, Y.~Guo, and J.~Ye, ``{Object Detection in 20 Years: A
  Survey},'' \emph{Proceedings of the IEEE}, 2023.

\bibitem{yolov5}
\BIBentryALTinterwordspacing
G.~Jocher, ``{YOLOv5 by Ultralytics},'' 2020. [Online]. Available:
  \url{https://github.com/ultralytics/yolov5}
\BIBentrySTDinterwordspacing

\bibitem{fcos}
Z.~Tian, C.~Shen, H.~Chen, and T.~He, ``{FCOS: Fully Convolutional One-Stage
  Object Detection},'' in \emph{Proceedings of the IEEE International
  Conference on Computer Vision (ICCV)}, 2019.

\bibitem{fastrcnn}
R.~Girshick, ``{Fast R-CNN},'' in \emph{Proceedings of the IEEE International
  Conference on Computer Vision (ICCV)}, 2015.

\bibitem{csppedet}
W.~Liu, I.~Hasan, and S.~Liao, ``{Center and Scale Prediction: Anchor-Free
  Approach for Pedestrian and Face Detection},'' \emph{Pattern Recognition},
  2023.

\bibitem{generalizablepedet}
I.~Hasan, S.~Liao, J.~Li, S.~U. Akram, and L.~Shao, ``{Generalizable Pedestrian
  Detection: The Elephant in the Room},'' in \emph{Proceedings of the IEEE
  Conference on Computer Vision and Pattern Recognition (CVPR)}, 2021.

\bibitem{illuminationaware}
C.~Li, D.~Song, R.~Tong, and M.~Tang, ``{Illumination-Aware Faster R-CNN for
  Robust Multispectral Pedestrian Detection},'' \emph{Pattern Recognition},
  2019.

\bibitem{RITA}
Y.~Liu, C.~Hu, B.~Zhao, Y.~Huang, and X.~Zhang, ``{Region-Based
  Illumination-Temperature Awareness and Cross-Modality Enhancement for
  Multispectral Pedestrian Detection},'' \emph{IEEE Transactions on Intelligent
  Vehicles}, pp. 1--12, 2024.

\bibitem{adaptation}
Q.~Xie, T.-Y. Cheng, Z.~Dai, V.~Tran, N.~Trigoni, and A.~Markham,
  ``{Illumination-Aware Hallucination-Based Domain Adaptation for Thermal
  Pedestrian Detection},'' \emph{IEEE Transactions on Intelligent
  Transportation Systems}, 2023.

\bibitem{IGT}
K.~Chen, J.~Liu, and H.~Zhang, ``{IGT: Illumination-Guided RGB-T Object
  Detection with Transformers},'' \emph{Knowledge-Based Systems}, vol. 268, p.
  110423, 2023.

\bibitem{TINet}
Y.~Zhang, H.~Yu, Y.~He, X.~Wang, and W.~Yang, ``{Illumination-Guided RGBT
  Object Detection with Inter- and Intra-Modality Fusion},'' \emph{IEEE
  Transactions on Instrumentation and Measurement}, 2023.

\bibitem{TIRDet}
Z.~Wang, F.~Colonnier, J.~Zheng, J.~Acharya, W.~Jiang, and K.~Huang, ``{TIRDet:
  Mono-Modality Thermal InfraRed Object Detection Based on Prior
  Thermal-To-Visible Translation},'' in \emph{Proceedings of the ACM
  International Conference on Multimedia (ACM MM)}, 2023, p. 2663–2672.

\bibitem{M2FNet}
X.~Li, S.~Chen, C.~Tian, H.~Zhou, and Z.~Zhang, ``{M2FNet: Mask-guided
  Multi-level Fusion for RGB-T Pedestrian Detection},'' \emph{IEEE Transactions
  on Multimedia}, pp. 1--13, 2024.

\bibitem{CAT+MFT}
W.-Y. Lee, L.~Jovanov, and W.~Philips, ``{Cross-Modality Attention and
  Multimodal Fusion Transformer for Pedestrian Detection},'' in
  \emph{Proceedings of the European Conference on Computer Vision (ECCV)},
  2022, pp. 608--623.

\bibitem{MSAT}
S.~You, X.~Xie, Y.~Feng, C.~Mei, and Y.~Ji, ``{Multi-Scale Aggregation
  Transformers for Multispectral Object Detection},'' \emph{IEEE Signal
  Processing Letters}, 2023.

\bibitem{ARCNN1}
L.~Zhang, Z.~Liu, X.~Zhu, Z.~Song, X.~Yang, Z.~Lei, and H.~Qiao, ``{Weakly
  Aligned Feature Fusion for Multimodal Object Detection},'' \emph{IEEE
  Transactions on Neural Networks and Learning Systems}, pp. 1--15, 2021.

\bibitem{C2Former}
M.~Yuan and X.~Wei, ``{C²Former: Calibrated and Complementary Transformer for
  RGB-Infrared Object Detection},'' \emph{IEEE Transactions on Geoscience and
  Remote Sensing}, vol.~62, pp. 1--12, 2024.

\bibitem{zhang2021weakly}
D.~Zhang, J.~Han, G.~Cheng, and M.-H. Yang, ``{Weakly Supervised Object
  Localization and Detection: A Survey},'' \emph{IEEE Transactions on Pattern
  Analysis and Machine Intelligence}, vol.~44, no.~9, pp. 5866--5885, 2021.

\bibitem{gaff}
H.~Zhang, E.~Fromont, S.~Lef{\`e}vre, and B.~Avignon, ``{Guided Attentive
  Feature Fusion for Multispectral Pedestrian Detection},'' in
  \emph{Proceedings of the IEEE Winter Conference on Applications of Computer
  Vision (WACV)}, 2021.

\bibitem{tsne}
L.~Van~der Maaten and G.~Hinton, ``{Visualizing Data Using t-SNE},''
  \emph{Journal of Machine Learning Research}, 2008.

\bibitem{fpn}
T.-Y. Lin, P.~Dollar, R.~Girshick, K.~He, B.~Hariharan, and S.~Belongie,
  ``{Feature Pyramid Networks for Object Detection},'' in \emph{Proceedings of
  the IEEE Conference on Computer Vision and Pattern Recognition (CVPR)}, 2017.

\bibitem{panet}
S.~Liu, L.~Qi, H.~Qin, J.~Shi, and J.~Jia, ``{Path Aggregation Network for
  Instance Segmentation},'' in \emph{Proceedings of the IEEE Conference on
  Computer Vision and Pattern Recognition (CVPR)}, 2018.

\bibitem{dfanet}
R.~Zhang, L.~Li, Q.~Zhang, J.~Zhang, L.~Xu, B.~Zhang, and B.~Wang,
  ``{Differential Feature Awareness Network within Antagonistic Learning for
  Infrared-Visible Object Detection},'' \emph{IEEE Transactions on Circuits and
  Systems for Video Technology}, 2023.

\bibitem{dcnv2}
X.~Zhu, H.~Hu, S.~Lin, and J.~Dai, ``{Deformable ConvNets v2: More Deformable,
  Better Results},'' in \emph{Proceedings of the IEEE Conference on Computer
  Vision and Pattern Recognition (CVPR)}, 2019.

\bibitem{focusnet}
X.~Zhang, Z.~Sheng, and H.-L. Shen, ``{FocusNet: Classifying Better by Focusing
  on Confusing Classes},'' \emph{Pattern Recognition}, 2022.

\bibitem{diceloss}
F.~Milletari, N.~Navab, and S.-A. Ahmadi, ``{V-Net: Fully Convolutional Neural
  Networks for Volumetric Medical Image Segmentation},'' in \emph{International
  Conference on 3D Vision (3DV)}, 2016.

\bibitem{vgg}
K.~Simonyan and A.~Zisserman, ``{Very Deep Convolutional Networks for
  Large-Scale Image Recognition},'' in \emph{International Conference on
  Learning Representations (ICLR)}, 2015.

\bibitem{DeformableDETR}
X.~Zhu, W.~Su, L.~Lu, B.~Li, X.~Wang, and J.~Dai, ``{Deformable DETR:
  Deformable Transformers for End-to-End Object Detection},'' in
  \emph{Proceedings of International Conference on Learning Representations
  (CVPR)}, 2021.

\bibitem{DINO}
H.~Zhang, F.~Li, S.~Liu, L.~Zhang, H.~Su, J.~Zhu, L.~Ni, and H.-Y. Shum,
  ``{DINO: DETR with Improved DeNoising Anchor Boxes for End-to-End Object
  Detection},'' in \emph{Proceedings of International Conference on Learning
  Representations (ICLR)}, 2023.

\bibitem{GUAN2019148}
D.~Guan, Y.~Cao, J.~Yang, Y.~Cao, and M.~Y. Yang, ``{Fusion of Multispectral
  Data Through Illumination-Aware Deep Neural Networks for Pedestrian
  Detection},'' \emph{Information Fusion}, 2019.

\bibitem{cian}
L.~Zhang, Z.~Liu, S.~Zhang, X.~Yang, H.~Qiao, K.~Huang, and A.~Hussain,
  ``{Cross-Modality Interactive Attention Network for Multispectral Pedestrian
  Detection},'' \emph{Information Fusion}, 2019.

\bibitem{msr}
J.~U. Kim, S.~Park, and Y.~M. Ro, ``{Towards Versatile Pedestrian Detector with
  Multisensory-Matching and Multispectral Recalling Memory},''
  \emph{{Proceedings of the AAAI Conference on Artificial Intelligence
  (AAAI)}}, 2022.

\bibitem{CMM}
T.~Kim, S.~Shin, Y.~Yu, H.~G. Kim, and Y.~M. Ro, ``{Causal Mode Multiplexer: A
  Novel Framework for Unbiased Multispectral Pedestrian Detection},'' in
  \emph{Proceedings of the IEEE Conference on Computer Vision and Pattern
  Recognition (CVPR)}, 2024.

\bibitem{cmpd}
Q.~Li, C.~Zhang, Q.~Hu, H.~Fu, and P.~Zhu, ``{Confidence-Aware Fusion using
  Dempster-Shafer Theory for Multispectral Pedestrian Detection},'' \emph{IEEE
  Transactions on Multimedia}, 2022.

\bibitem{mlpd}
J.~Kim, H.~Kim, T.~Kim, N.~Kim, and Y.~Choi, ``{MLPD: Multi-Label Pedestrian
  Detector in Multispectral Domain},'' \emph{IEEE Robotics and Automation
  Letters}, 2021.

\bibitem{SMPD}
Q.~Li, C.~Zhang, Q.~Hu, P.~Zhu, H.~Fu, and L.~Chen, ``{Stabilizing
  Multispectral Pedestrian Detection with Evidential Hybrid Fusion},''
  \emph{IEEE Transactions on Circuits and Systems for Video Technology}, 2023.

\bibitem{MCHE-CF}
R.~Li, J.~Xiang, F.~Sun, Y.~Yuan, L.~Yuan, and S.~Gou, ``{Multiscale
  Cross-Modal Homogeneity Enhancement and Confidence-Aware Fusion for
  Multispectral Pedestrian Detection},'' \emph{IEEE Transactions on
  Multimedia}, vol.~26, pp. 852--863, 2024.

\bibitem{CPFM}
C.~Tian, Z.~Zhou, Y.~Huang, G.~Li, and Z.~He, ``{Cross-Modality Proposal-Guided
  Feature Mining for Unregistered RGB-Thermal Pedestrian Detection},''
  \emph{IEEE Transactions on Multimedia}, vol.~26, 2024.

\bibitem{vtfyolo}
X.~Zou, T.~Peng, and Y.~Zhou, ``{UAV-Based Human Detection With Visible-Thermal
  Fused YOLOv5 Network},'' \emph{IEEE Transactions on Industrial Informatics},
  vol.~20, no.~3, pp. 3814--3823, 2024.

\bibitem{AANet}
N.~Chen, J.~Xie, J.~Nie, J.~Cao, Z.~Shao, and Y.~Pang, ``{Attentive Alignment
  Network for Multispectral Pedestrian Detection},'' in \emph{Proceedings of
  ACM International Conference on Multimedia (ACM MM)}, 2023, p. 3787–3795.

\bibitem{YOLO-Adaptor}
H.~Fu, H.~Liu, J.~Yuan, X.~He, J.~Lin, and Z.~Li, ``{YOLO-Adaptor: A Fast
  Adaptive One-Stage Detector for Non-Aligned Visible-Infrared Object
  Detection},'' \emph{IEEE Transactions on Intelligent Vehicles}, pp. 1--14,
  2024.

\bibitem{AlignedFLIR}
H.~Zhang, E.~Fromont, S.~Lefevre, and B.~Avignon, ``{Multispectral Fusion for
  Object Detection with Cyclic Fuse-and-Refine Blocks},'' in \emph{Proceedings
  of International Conference on Image Processing (ICIP)}, 2020, pp. 276--280.

\bibitem{Fusion-Mamba}
W.~Dong, H.~Zhu, S.~Lin, X.~Luo, Y.~Shen, X.~Liu, J.~Zhang, G.~Guo, and
  B.~Zhang, ``{Fusion-Mamba for Cross-Modality Object Detection},'' \emph{arXiv
  preprint arXiv:2404.09146}, 2024.

\bibitem{EME}
X.~Zhang, S.-Y. Cao, F.~Wang, R.~Zhang, Z.~Wu, X.~Zhang, X.~Bai, and H.-L.
  Shen, ``{Rethinking Early-Fusion Strategies for Improved Multispectral Object
  Detection},'' \emph{arXiv preprint arXiv:2405.16038}, 2024.

\bibitem{mmdetection}
\BIBentryALTinterwordspacing
{MMDetection Contributors}, ``{OpenMMLab Detection Toolbox and Benchmark},''
  2018. [Online]. Available: \url{https://github.com/open-mmlab/mmdetection}
\BIBentrySTDinterwordspacing

\bibitem{piotr}
P.~Dollar, C.~Wojek, B.~Schiele, and P.~Perona, ``{Pedestrian Detection: An
  Evaluation of the State of the Art},'' \emph{IEEE Transactions on Pattern
  Analysis and Machine Intelligence}, 2012.

\bibitem{resnet}
K.~He, X.~Zhang, S.~Ren, and J.~Sun, ``{Deep Residual Learning for Image
  Recognition},'' in \emph{Proceedings of the IEEE Conference on Computer
  Vision and Pattern Recognition (CVPR)}, 2016.

\bibitem{detfusion}
Y.~Sun, B.~Cao, P.~Zhu, and Q.~Hu, ``{DetFusion: A Detection-Driven Infrared
  and Visible Image Fusion Network},'' in \emph{Proceedings of the ACM
  International Conference on Multimedia (ACM MM)}, 2022.

\bibitem{swin}
Z.~Liu, Y.~Lin, Y.~Cao, H.~Hu, Y.~Wei, Z.~Zhang, S.~Lin, and B.~Guo, ``{Swin
  Transformer: Hierarchical Vision Transformer using Shifted Windows},'' in
  \emph{Proceedings of the International Conference on Computer Vision (ICCV)},
  2021.

\bibitem{IEGOD}
H.~Liu, F.~Jin, H.~Zeng, H.~Pu, and B.~Fan, ``{Image Enhancement Guided Object
  Detection in Visually Degraded Scenes},'' \emph{IEEE Transactions on Neural
  Networks and Learning Systems}, pp. 1--14, 2023.

\bibitem{HalluciDet}
H.~R. Medeiros, F.~A. Guerrero~Peña, M.~Aminbeidokhti, T.~Dubail, E.~Granger,
  and M.~Pedersoli, ``{HalluciDet: Hallucinating RGB Modality for Person
  Detection Through Privileged Information},'' in \emph{Proceedings of the
  Winter Conference on Applications of Computer Vision (WACV)}, 2024, pp.
  1433--1442.

\bibitem{PoolFuser}
Y.~Cao, Y.~Fan, J.~Bin, and Z.~Liu, ``{Lightweight Transformer for Multi-Modal
  Object Detection (Student Abstract)},'' in \emph{Proceedings of the AAAI
  Conference on Artificial Intelligence (AAAI)}, vol.~37, no.~13, 2023, pp.
  16\,172--16\,173.

\bibitem{DIVFusion}
L.~Tang, X.~Xiang, H.~Zhang, M.~Gong, and J.~Ma, ``{DIVFusion: Darkness-Free
  Infrared and Visible Image Fusion},'' \emph{Information Fusion}, vol.~91, pp.
  477--493, 2023.

\bibitem{LENFusion}
J.~Chen, L.~Yang, W.~Liu, X.~Tian, and J.~Ma, ``{LENFusion: A Joint Low-Light
  Enhancement and Fusion Network for Nighttime Infrared and Visible Image
  Fusion},'' \emph{IEEE Transactions on Instrumentation and Measurement}, 2024.

\bibitem{DM-Fusion}
G.~Xu, C.~He, H.~Wang, H.~Zhu, and W.~Ding, ``{DM-Fusion: Deep Model-Driven
  Network for Heterogeneous Image Fusion},'' \emph{IEEE Transactions on Neural
  Networks and Learning Systems}, pp. 1--15, 2023.

\bibitem{CAMF}
L.~Tang, Z.~Chen, J.~Huang, and J.~Ma, ``{CAMF: An Interpretable Infrared and
  Visible Image Fusion Network Based on Class Activation Mapping},'' \emph{IEEE
  Transactions on Multimedia}, 2023.

\bibitem{MoE-Fusion}
B.~Cao, Y.~Sun, P.~Zhu, and Q.~Hu, ``{Multi-Modal Gated Mixture of
  Local-to-Global Experts for Dynamic Image Fusion},'' in \emph{Proceedings of
  the International Conference on Computer Vision (ICCV)}, 2023, pp.
  23\,555--23\,564.

\bibitem{MetaFusion}
W.~Zhao, S.~Xie, F.~Zhao, Y.~He, and H.~Lu, ``{MetaFusion: Infrared and Visible
  Image Fusion via Meta-Feature Embedding from Object Detection},'' in
  \emph{Proceedings of the Conference on Computer Vision and Pattern
  Recognition (CVPR)}, 2023, pp. 13\,955--13\,965.

\bibitem{TFNet}
F.~Chu, J.~Cao, Z.~Song, Z.~Shao, Y.~Pang, and X.~Li, ``{Toward Generalizable
  Multispectral Pedestrian Detection},'' \emph{IEEE Transactions on Intelligent
  Transportation Systems}, 2023.

\bibitem{DDFM}
Z.~Zhao, H.~Bai, Y.~Zhu, J.~Zhang, S.~Xu, Y.~Zhang, K.~Zhang, D.~Meng,
  R.~Timofte, and L.~Van~Gool, ``{DDFM: Denoising Diffusion Model for
  Multi-Modality Image Fusion},'' in \emph{Proceedings of the International
  Conference on Computer Vision (ICCV)}, 2023, pp. 8082--8093.

\bibitem{CSAA}
Y.~Cao, J.~Bin, J.~Hamari, E.~Blasch, and Z.~Liu, ``{Multimodal Object
  Detection by Channel Switching and Spatial Attention},'' in \emph{Proceedings
  of the Conference on Computer Vision and Pattern Recognition (CVPR)}, 2023,
  pp. 403--411.

\bibitem{Diff-IF}
X.~Yi, L.~Tang, H.~Zhang, H.~Xu, and J.~Ma, ``{Diff-IF: Multi-Modality Image
  Fusion via Diffusion Model with Fusion Knowledge Prior},'' \emph{Information
  Fusion}, p. 102450, 2024.

\bibitem{CALNet}
X.~He, C.~Tang, X.~Zou, and W.~Zhang, ``{Multispectral Object Detection via
  Cross-Modal Conflict-Aware Learning},'' in \emph{Proceedings of the ACM
  International Conference on Multimedia (ACM MM)}, 2023, pp. 1465--1474.

\bibitem{EGMT}
D.-G. Lee, M.-H. Jeon, Y.~Cho, and A.~Kim, ``{Edge-Guided Multi-Domain
  RGB-to-TIR Image Translation for Training Vision Tasks with Challenging
  Labels},'' in \emph{Proceedings of the International Conference on Robotics
  and Automation (ICRA)}.\hskip 1em plus 0.5em minus 0.4em\relax IEEE, 2023,
  pp. 8291--8298.

\bibitem{CE-RetinaNet}
Y.~Zhang and Z.~Cai, ``{CE-RetinaNet: A Channel Enhancement Method for Infrared
  Wildlife Detection in UAV Images},'' \emph{IEEE Transactions on Geoscience
  and Remote Sensing}, 2023.

\bibitem{ICAFusion}
J.~Shen, Y.~Chen, Y.~Liu, X.~Zuo, H.~Fan, and W.~Yang, ``{ICAFusion: Iterative
  Cross-Attention Guided Feature Fusion for Multispectral Object Detection},''
  \emph{Pattern Recognition}, vol. 145, p. 109913, 2024.

\bibitem{PSFusion}
L.~Tang, H.~Zhang, H.~Xu, and J.~Ma, ``{Rethinking the Necessity of Image
  Fusion in High-Level Vision Tasks: A Practical Infrared and Visible Image
  Fusion Network Based on Progressive Semantic Injection and Scene Fidelity},''
  \emph{Information Fusion}, vol.~99, p. 101870, 2023.

\bibitem{AUIF}
Z.~Zhao, S.~Xu, J.~Zhang, C.~Liang, C.~Zhang, and J.~Liu, ``{Efficient and
  Model-Based Infrared and Visible Image Fusion via Algorithm Unrolling},''
  \emph{IEEE Transactions on Circuits and Systems for Video Technology},
  vol.~32, no.~3, pp. 1186--1196, 2022.

\bibitem{CDDFuse}
Z.~Zhao, H.~Bai, J.~Zhang, Y.~Zhang, S.~Xu, Z.~Lin, R.~Timofte, and
  L.~Van~Gool, ``{CDDFuse: Correlation-Driven Dual-Branch Feature Decomposition
  for Multi-Modality Image Fusion},'' in \emph{Proceedings of the Conference on
  Computer Vision and Pattern Recognition (CVPR)}, 2023, pp. 5906--5916.

\bibitem{U2Fusion}
H.~Xu, J.~Ma, J.~Jiang, X.~Guo, and H.~Ling, ``{U2Fusion: A Unified
  Unsupervised Image Fusion Network},'' \emph{IEEE Transactions on Pattern
  Analysis and Machine Intelligence}, vol.~44, no.~1, pp. 502--518, 2020.

\end{thebibliography}

\end{document}